\RequirePackage{fix-cm}
\documentclass[twocolumn]{svjour3}
\smartqed  % flush right qed marks, e.g. at end of proof
\usepackage{graphicx}
\usepackage{booktabs}
\usepackage{colortbl}
\usepackage{makecell}
\usepackage{tikz,pgfplots,pgfplotstable}
\usepackage{color}
\usepackage{amsfonts}
\usepackage{amssymb}
\usepackage{multirow}
\usepackage{tcolorbox}
\usepackage{marvosym} 
\usepackage{algorithm}
\usepackage{newfloat}
\usepackage{listings}
\usepackage{xcolor}
\usepackage{amsmath}
\usepackage{bm}
\usepackage{algpseudocode} 

\usepgfplotslibrary[groupplots]

\usepackage[pagebackref=true,breaklinks=true,colorlinks,bookmarks=false]{hyperref}

\newcommand{\ours}{SAGE}

\definecolor{mypink1}{RGB}{241,90,82}
\definecolor{mypink}{RGB}{255,231,226}
\definecolor{myblue}{RGB}{233,250,249}

\definecolor{mygray}{gray}{0.6}
\newcommand{\pub}[1]{\color{mygray}{\scriptsize{[{#1}]}}}

\newcommand{\ie}{\textit{i}.\textit{e}.}
\newcommand{\eg}{\textit{e}.\textit{g}.}

\begin{document}

\title{SAGE: Exploring the Boundaries of Unsafe Concept Domain with Semantic-Augment Erasing}

\author{Hongguang Zhu \and Yunchao Wei \and Mengyu Wang \and Siyu Jiao \and Yan Fang \and Jiannan Huang \and Yao Zhao$^{(\textrm{\Letter})}$}

% \authorrunning{Short form of author list} % if too long for running head
\institute{
{{\textbullet{} Hongguang Zhu is with Faculty of Data Science, City University of Macau, \email{\{zhuhongguang1103\}@gmail.com};\\}
{\textbullet{} Yunchao Wei, Mengyu Wang, Siyu Jiao, Yan Fang, Jiannan Huang, Yao Zhao are with Institute of Information Science, Beijing Jiaotong University, and Beijing Key Laboratory of Advanced Information Science and Network Technology;\\}
{\textbullet{} Yao Zhao is the corresponding author of this work, \email{\{yzhao\}@bjtu.edu.cn};}
}
}

% \institute{
% Hongguang Zhu \at 
% Faculty of Data Science, City University of Macau. \\
% \email{\{zhuhongguang1103\}@gmail.com}.
% \and
% Yunchao Wei, Mengyu Wang, Jiannan Huang，Siyu Jiao, Yan Fang, Yao Zhao \at
% Institute of Information Science, Beijing Jiaotong University. \\
% Beijing Key Laboratory of Advanced Information Science and Network Technology. \\
% \email{\{wychao1987,wmengyu0826，jiannan2003\}@gmail.com}, {\{jiaosiyu99,yanfang,yzhao\}@bjtu.edu.cn}\\
% }

\date{Received: date / Accepted: date}

\maketitle

\begin{abstract}
Diffusion models (DMs) have achieved significant progress in text-to-image generation. However, the inevitable inclusion of sensitive information during pre-training poses safety risks, such as unsafe content generation and copyright infringement. Concept erasing finetunes weights to unlearn undesirable concepts, and has emerged as a promising solution.
However, existing methods treat unsafe concept as a fixed word and repeatedly erase it, trapping DMs in ``word concept abyss'', which prevents generalized concept-related erasing.
To escape this abyss, we introduce \textbf{semantic-augment erasing} which transforms concept word erasure into concept domain erasure by the cyclic self-check and self-erasure.
It efficiently explores and unlearns the boundary representation of concept domain through semantic spatial relationships between original and training DMs, without requiring additional preprocessed data.
Meanwhile, to mitigate the retention degradation of irrelevant concepts while erasing unsafe concepts, we further propose the \textbf{global-local collaborative retention} mechanism that combines global semantic relationship alignment with local predicted noise preservation, effectively expanding the retentive receptive field for irrelevant concepts. 
We name our method SAGE, and extensive experiments demonstrate the comprehensive superiority of SAGE compared with other methods in the safe generation of DMs. 
The code and weights will be open-sourced at \url{https://github.com/KevinLight831/SAGE}.

\noindent\textit{\textbf{Warning}:{
This paper contains potentially offensive outputs.}}

\end{abstract}

\section{Introduction}
\label{sec:intro}
\begin{figure*}[!tbp]
	\centering
    \includegraphics[width=1\textwidth]{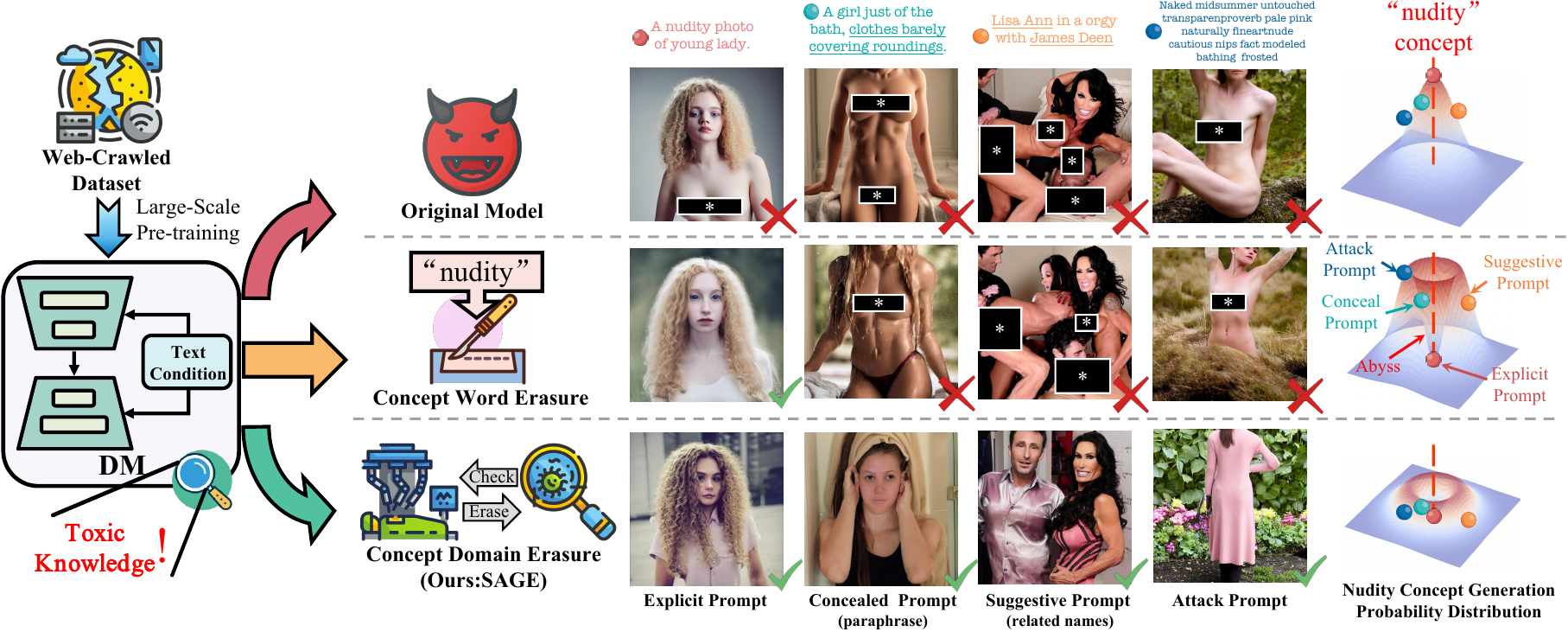} 
	\caption{
Pre-trained diffusion models (DMs) inevitably memorize toxic knowledge, leading to unsafe content generation issues. Previous concept erasure methods are trapped in 
 word concept abyss by repeatedly erasing specific word (\eg, \texttt{nudity}), failing on concept-related prompts, \ie, concealed prompts, suggestive prompts (\eg, \texttt{porn star names}) and attack prompts. Our {\ours} proposes concept domain erasure through cyclic self-check and self-erasure. It can efficiently achieve the model self-purification while preserving model utility. \textit{The \fbox{black boxs} with $\ast$ are added by authors for publication.}
 }
	\label{fig:intro}
\end{figure*}
Recent advancements in text-to-image diffusion models \cite{dhariwal2021diffusion,SD1_4,ramesh2022hierarchical,ding2022cogview2} have led to significant achievements in producing photo-realistic images, proving beneficial for various industrial applications \cite{midjourney,dalle2,animatediff,sv3d}. 
However, due to the extensive use of web-scraped datasets during training, these models pose significant challenges, including the generation of unsafe content (\ie, erotic, violent, drug, illegal) \cite{tatum2023porn,Hayden2024} and the replication of copyrighted material \cite{setty2023suit,jiang2023ai,roose2022art}. 
One intuitive solution is to filter inappropriate images and retrain DMs manually. However, this approach \cite{rombach2022sd2} is not only computationally expensive but also potentially incomplete erasure \cite{esd}.
Additionally, using Safety Checkers \cite{rando2022red} to detect and mitigate harmful outputs offers an alternative, but this approach depends on the accuracy of the detector and is limited by intrinsic biases.
 
In response to the above challenges, concept erasing \cite{esd} has emerged as a potentially promising solution. Specifically, given a concept described in text, the pre-trained model is fine-tuned to forget the related memory of that concept, thus preventing the generation of associated content.
Some approaches use preprocessed pairs of images and masks \cite{liu2023grounding} to suppress attention activation \cite{MACE,receler} in concept-related regions, or remap the target concept to a benign one \cite{UCE,RECE} using pre-defined pairs of target and benign prompts.
However, these methods have notable limitations in generalizing erasure.
(1) \textbf{Explicit Concept Representation}: they rigidly define the concept as a specific word (\eg, \texttt{nudity}) and repeatedly erase it, leading to erasure overfitting on single word. (2) \textbf{Explicit Erasure Mode}: whether through suppressing attention regions or remapping concept words, these methods are constrained by the knowledge within the preprocessed data and cannot harness the inherent knowledge of model for implicit self-erasure.
As illustrated in Fig. \ref{fig:intro}, these limitations result in post-erasure DMs still generating unsafe content when confronted with concealed prompts, suggestive prompts, and attack prompts. 

Concept-related descriptions are diverse and cannot be exhaustively listed, but large models inherently contain compressed world knowledge \cite{zhang2023large}. 
Therefore, it is more cost-effective and promising to perform concept surgery (erasure or modification) leveraging the knowledge topology within the large model itself. In other words, 
\textbf{\textit{the one who opens the Pandora's box must close it}.}
To achieve this, we transform traditional fixed-word concept representations into self-augment token embeddings. Unlike static discrete words, token embedding can be continuously updated based on the training feedback from DM.
By freezing the denoiser responsible for image generation and adjusting the text encoder that manages conditional mapping, we establish a connection between the textual space of current training DM and the visual space of original freeze DM. 
Further, the self-augment token embedding can be efficiently optimized to explore the boundaries of the target concept domain. 
By aligning the opposite direction of guidance noise produced by the optimized self-augment token embedding, DM can steer the generation tendency away from the target concept domain, facilitating self-purification.
Excessive erasure of unsafe concepts will inevitably degrade the retention of irrelevant concepts, potentially affecting the usability of concept-erased DMs in severe cases. 
To this end, we propose a global-local collaborative retention mechanism for irrelevant concepts. It first aligns the textual semantic graph of irrelevant concepts at the global relational level, then identifies concepts with the most significant semantic drift to apply additional local predictive noise retention constraints. 
We call our approach \textbf{S}emantic-\textbf{A}ugment concept erasing with \textbf{G}lobal-local collaborative r\textbf{E}tention (\textbf{\ours}).
Extensive experiments demonstrate that {\ours} achieves advanced comprehensive performance in erasing target concepts while preserving non-target concepts. Moreover, our method has high training efficiency and supports zero-cost migration within the same series of DMs.

\section{Related Work}
\label{sec:related}
\noindent\textbf{Safe generation of DMs.}
Leveraging training on large-scale web-crawled datasets, DMs \cite{SD1_4,sd3,sdxl} can generate high-quality images and exhibit immense creative potential. However, since these datasets are not curated, DMs inadvertently memorize unsafe and copyrighted content, leading to unsafe generation. To mitigate this issue, the efforts can be classified into three aspects: (1) \textit{Pre-processing}, (2) \textit{Post-processing}, and (3) \textit{Model editing}. 

\noindent\textbf{\textit{{Pre-processing methods}}} utilize pretrained detectors to filter out images containing unsafe content, and retrain DMs after filtering. However, retraining from scratch is computationally expensive and impractical for addressing evolving erasure requests. \eg, Stable Diffusion v2.1 \cite{rombach2022sd2} consumes 150,000 GPU hours to retrain on the filtered LAION-5B dataset \cite{schuhmann2022laion}. This extensive filtering process also has been found to negatively impact output quality \cite {connor2022sd}, and DMs may still not be properly sanitized \cite{esd}.

\noindent\textbf{\textit{Post-processing methods}} use safety checkers to identify unsafe content and block risky outputs. 
Several organizations \cite{midjourney,dalle2} deploy this approach by involving a blacklist-style post-hoc filter. However, safety checkers of open-source models can be easily circumvented by modifying code \cite{SmithMano2022}. The safety filters of closed-source models like DALL$\cdot$E 2 \cite{dalle2} can be bypassed using attack prompts \cite{sneakyprompt,RingABell}. Similar to pre-processing methods, Post-processing methods also rely on the accuracy of detectors, whose inherent biases can result in unreliable exclusion of unsafe content. 

\noindent\textbf{\textit{Model editing methods}} leverage the original DMs to erase or redirect target concepts, effectively eliminating potential harmful biases before deployment. Due to their low cost, flexible operation, and effective erasure, these methods have increasingly attracted community interest.
We categorize existing model editing methods for concept erasing into three categories based on their technical characteristics.

\noindent\textbf{I. \textit{Guidance-based Methods}} \cite{SLD,sdd,SPM,AC,esd}: 
SLD \cite{SLD} modifies denoising process in inference stage and introduces negative guidance to prevent unsafe content generation. However, it only suppresses undesired concepts in inference rather than complete removal.
ESD \cite{esd} predicts negative guided noises and trains DM to steer conditional predictions away from target concepts. 

\noindent\textbf{II. \textit{Attention Re-steering Methods}} \cite{FMN,MACE} employ attention re-steering to identify regions associated with target concepts within the cross-attention layers of UNet \cite{ronneberger2015u,ddpm}. By diminishing the cross-attention activation related to target concepts, DMs gradually disregard these concepts during image generation. However, the effectiveness of this method depends heavily on the accurate location of the concept-related region and the quality of pre-processing images. Besides, this method is limited to scenarios where the concept-related regions can be explicitly identified. For strongly coupled concepts like artistic styles, it is difficult to accurately pre-process masks that isolate content purely related to style without capturing object-specific details.

\noindent\textbf{III. \textit{Closed-form Editing Methods}} \cite{UCE,MACE,RECE} optimize the key and value projection matrices in the cross-attention layers of UNet. Specifically, UCE \cite{UCE} recalibrates the embedding of a target prompt (\eg, \texttt{nudity}) to a benign prompt (\eg, \texttt{wearing clothes}), while keeping other concepts unchanged. 
MACE \cite{MACE} first erases a single prompt using the attention re-steering method and then jointly optimizes the projection matrices. Despite a certain effectiveness, they cannot achieve generalized erasure against concealed prompts, suggestive prompts, and attack prompts \cite{RingABell,sneakyprompt}. Since concept-related prompts are inexhaustible, these methods, which focus only on human-cognizable language specified in the dataset, fail to address the broader spectrum of machine-cognizable language.

Recent red-teaming works \cite{zhang2024adversarial,RingABell,prompting4debugging} have leveraged the idea of textual inversion \cite{gal2022image} to generate attack prompts that provoke concept-erased DMs to regenerate unsafe images. 
Inspired by this, some works try to introduce red-teaming methods \cite{prompting4debugging,zhang2023generate} to generate attack prompts and further support adversarial training to improve robustness.
RACE \cite{RACE} optimize random perturbation into attack perturbation to adversarially finetune UNet.
Receler \cite{receler} integrates a lightweight adapter within cross-attention layers, utilizing adversarial prompt learning to improve robustness.
RECE \cite{RECE} extends closed-form editing methods by incorporating adversarial fine-tuning on matrix-modified cross-attention layers.
AdvUnlearn \cite{Advunlearn} formulates the concept erasure as a bilevel optimization problem, simultaneously optimizing for both target concept removal and non-target concept preservation.
Our {\ours} leverages the modal space relationship between the current DM and the original DM to efficiently explore the boundaries of target concepts. Thus, there is no need for complex attack prompt optimization. Moreover, in contrast to most methods \cite{esd,RACE,UCE,MACE,receler,RECE} that modify the UNet through fine-tuning, our method only optimizes the text encoder. This design enables the purified text encoder to be directly deployed across DMs that share the same text encoder architecture, eliminating the need for retraining.

\section{Preliminary}
\label{sec:preliminary}
\noindent \textbf{Stable Diffusion Models.}
Our study builds upon Stable Diffusion (SD) Models \cite{SD1_4}, which incorporate conditional text prompts into image embedding to guide the generation process.
The diffusion process begins with a noise latent $\mathbf z$ drawn from a Gaussian distribution $\mathcal{N}(0, 1)$. Over a series of $T$ time steps, this noise latent undergoes a gradual denoising process guided by textual embedding \cite{clip}, transforming into a clean latent $\mathbf z_0$. 
Meanwhile, the encoder of pre-trained Variational Autoencoder (VAE) \cite{kingma2013auto,vqgan} transforms the input image $x$ into the latent $ \mathbf{z} =\mathcal{E}(x)$, and the decoder reconstructs the image from latent form, where $\mathcal{D}(\mathbf z)=\hat{x}\approx x$. Finally, the denoised latent is decoded into a clean image by decoder.
At each time step $t$, DM predicts noise using UNet denoiser ${\epsilon}_{\boldsymbol{\theta}}$,   parameterized by $\boldsymbol{\theta}$ and conditioned on the token embedding of input prompt $\tau$. The training objective for $\boldsymbol{\theta}$ is to minimize the denoising error, defined as:
\begin{equation}
\underset{\boldsymbol{\theta}}{\text{minimize}}~
\mathbb{E}_{\mathbf{z} \sim \mathcal{E}(x), \tau, t, {\epsilon} \sim \mathcal{N}(0,1)} \left[ \left\| {\epsilon}  - {\epsilon}_{\boldsymbol{\theta}}(\mathbf{z}_t, \tau, t) \right\|^2_2 \right],
\label{eq:ldm}
\end{equation}
where $\mathbf z_t$ is the noisy version of $\mathbf z$ up to the time step $t$. For notational simplicity, we will omit the time step $t$ in the following paragraphs.

\noindent \textbf{Concept Erasing in DM.}
Concept erasing was proposed to remove undesirable concepts from the latent space of DM. Inspired by classifier-free guidance \cite{ho2021classifier}, ESD \cite{esd} first proposes concept erasure by guiding the predicted noise away from the conditional noise of target concept $c$. The diffusion process of ESD can be denoted as:
\begin{equation}
{\epsilon}_{\boldsymbol{\theta}_\mathrm{n}}(\mathbf z_t,\tau_\mathrm{c}) \leftarrow
\underbrace{\epsilon_{{\boldsymbol{\theta}}_\mathrm{o}}(\mathbf z_t)-\eta[\epsilon_{{\boldsymbol{\theta}}_\mathrm{o}}(\mathbf z_t,\tau_\mathrm{c})-\epsilon_{{\boldsymbol{\theta}}_\mathrm{o}}(\mathbf z_t)]}_{\hat{\epsilon_{{\boldsymbol{\theta}}_\mathrm{n}}}(\mathbf z_t, \tau_\mathrm{c})},
\label{eq:erase_predict}
\end{equation}
where ${{\boldsymbol{\theta}}_\mathrm{o}}$ represents the original DM and ${\boldsymbol{\theta}_\mathrm{n}}$ denotes the training DM. $\tau_\mathrm{c}$ is the token embedding of concept word $c$, and $\eta$ denotes the guidance scale. 
$\epsilon_{{\boldsymbol{\theta}}_\mathrm{o}}(\mathbf z_t)$ is the noise predicted by the original DM with a null prompt input.
${\hat{\epsilon_{{\boldsymbol{\theta}}_\mathrm{n}}}(\mathbf z_t, \tau_\mathrm{c})}$ is the negative guidance noise. This process only needs the concept words to induce the intrinsic concept-related noise of DM. The erasure loss is formalized as:
\begin{align}
\mathcal{L}_{\mathrm{erase}} =\mathbb{E} \left[\left\|
\epsilon_{\boldsymbol{\theta}_\mathrm{n}}(\mathbf z_t,\tau_\mathrm{c}) - 
\hat{\epsilon_{{\boldsymbol{\theta}}_\mathrm{n}}}({\mathbf z_t,\tau_\mathrm{c}})
\right\|^2_2 \right], 
\label{eq:esd}
\end{align}

\noindent \textbf{Attack prompts against concept-erased DM.}
Red-teaming works \cite{prompting4debugging,zhang2023generate} seek to circumvent the erasure mechanisms and compel concept-erased DM to again generate harmful images using attack prompts. 
The token embedding of perturbed concept prompt, $\tau_\mathrm{c}^\prime$ is created by manipulating tokens or their embedding through random initialization \cite{RACE,receler,Advunlearn}.
The process for generating attack prompts can be represented as:
\begin{align}
\underset{\| \tau_\mathrm{c}^\prime - \tau_\mathrm{c}\| \leq \delta }{\text{minimize}}\quad&\mathbb{E} \left[ 
\left\| \epsilon_{\boldsymbol{\theta}_*}(\mathbf z_{t}, \tau_\mathrm{c}^\prime) - 
\epsilon_{{\boldsymbol{\theta}}_\mathrm{o}}(\mathbf z_{t}, \tau_\mathrm{c})
\right\|_2^2 
\right],
\label{eq:attack}
\end{align}
where ${\boldsymbol{\theta}_*}$ is the frozen victim concept-erased DM. The perturbed token embedding $\tau_\mathrm{c}^\prime$ is optimized by projected gradient descent (PGD) \cite{PGD} in the continuous textual embedding. Finally, Mapping the token embedding to the discrete texts to generate the attack prompt. 
Eq.\eqref{eq:attack} aims to optimize $\tau_\mathrm{c}^\prime$ that induce concept-erased DM ${\boldsymbol{\theta}_*}$ to regenerate unsafe content about $\tau_\mathrm{c}$.
The constraint in Eq.\eqref{eq:attack} ensures that $\tau_\mathrm{c}^\prime$ remains close to $\tau_\mathrm{c}$, subject to the added initial perturbation strength $\delta$.

\section{Method}
\label{sec:method}
\begin{figure*}[!tbp]
	\centering
	\includegraphics[width=\linewidth]{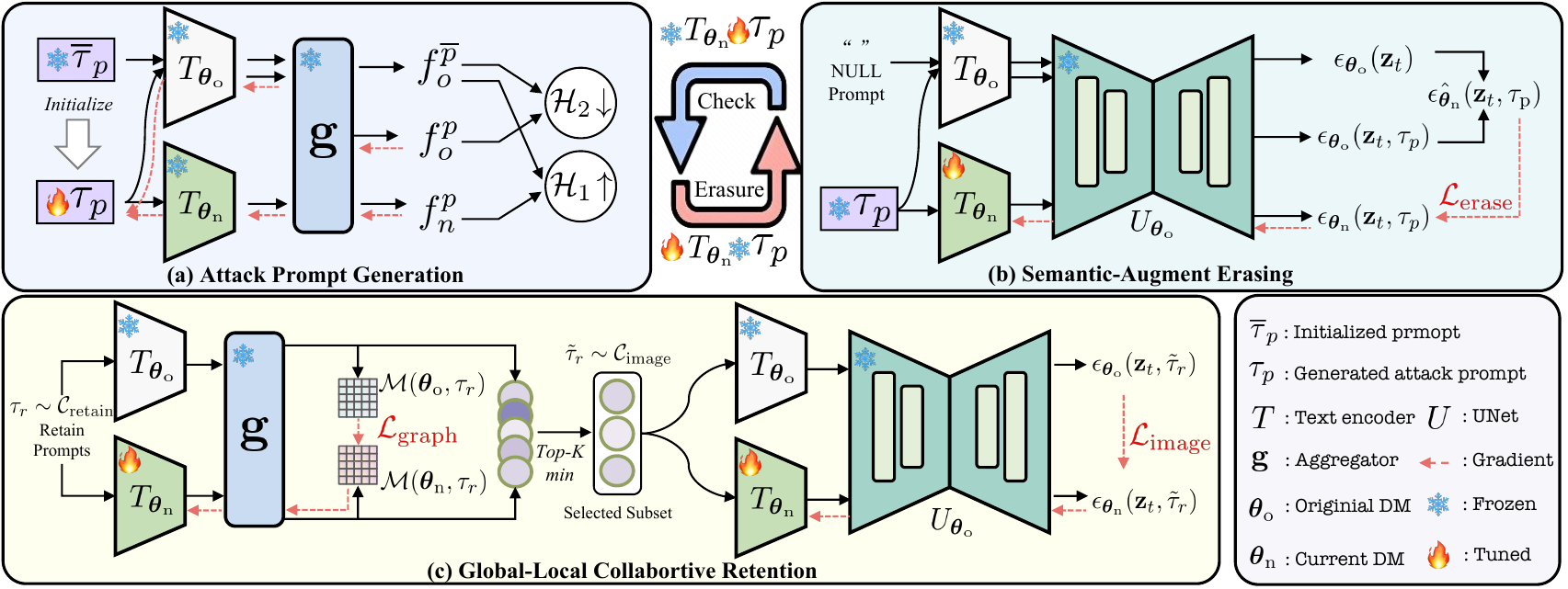}
	\vspace{-0.5cm}
	\caption{The proposed {\ours} involves three key components: the adversarial training of (a) attack prompt generation and (b) semantic-augment erasing, enabling the DM to self-check and self-erase; and (c) global-local collaborative retention mechanism that integrates semantic relations and predicted noise to preserve irrelevant concepts.
 }
	\vspace{-0.5cm}
	\label{fig:main}
\end{figure*}

We propose {\ours} to achieve concept domain erasing while preserving the native generation capability of DM. 
As depicted in Fig. \ref{fig:main}, our {\ours} consists of three main components: (a) \textit{Attack Prompt Generation}, (b) \textit{Semantic-Augment Erasing}, and (c) \textit{Global-Local Collaborative Retention}. 
The former two components generate attack prompts to further augment erasure based on the semantic space relationship.
The latter strives to maintain semantic alignment and generation capability of non-target concepts by simultaneously preserving global semantic relationships and local noise predictions. The following sections will provide a detailed explanation. 

\subsection{Semantic-Augment Erasing}
\label{sec:SAE}
To enable the DM to adaptively explore and erase the boundaries of the concept domain during training, our semantic-augment erasing operates in two phases:
attack prompt optimization and DM parameter optimization. 
These two phases continuously alternate, where each phase uses the optimized output of the other as input, thereby creating a cyclic adversarial training process of self-check and self-erasure.
Specifically, during attack prompt optimization, DM parameters remain fixed, and once the attack prompt is optimized, it is used as input to further finetune DM parameters.

The Eq.\eqref{eq:attack} describes a one-way attack optimization from the randomly perturbed prompt toward the target concept, just like what RACE \cite{RACE}, Receler \cite{receler}, and AdvUnlearn \cite{Advunlearn} conducted, which we refer to as the \textbf{outside-in methods}. 
The distribution randomness of their generated attack prompts in the concept domain depends on the randomness of initial perturbation $\delta$.
Besides, since the outside-in methods require a multi-round denoising process of the DM up to time $t$ for $\mathbf z_{t}$, and optimize $\delta$ from random initialization into the concept domain. It results in high computational and time consumption. 

In contrast, we propose the \textbf{inside-out methods} to explore the boundaries of concept domain starting from the target concept prompt rather than a perturbed prompt.
For example, when erasing the \texttt{Van Gogh} concept, attack prompt is first initialized by combining concept word with a randomly selected template from predefined template library (\eg `\texttt{An artwork by [Van Gogh]}'). 
For simplicity in formula form, the token embedding of original attack prompt can be denoted as $\tau_p = [\tau_{t}, \tau_c]$,  where $\tau_{t}$ and $\tau_c$ respectively represent the token embeddings of template and concept word. 
At this stage, the current training DM $\boldsymbol{\theta}_\mathrm{n}$ will be frozen and \textbf{only template token embedding $\bm{\tau_{t}}$ is optimized} to generate attack token embedding $\tau_p$ against $\boldsymbol{\theta}_\mathrm{n}$.
In other words, the original attack prompt is modeled as the centroid of target concept domain, while template token embedding is optimized to introduce semantic perturbations to attack prompt, thereby continuously exploring boundaries of target concept domain. 
Through such perturbations in the semantic space, it achieves efficient coverage of hard-to-quantify expressions (\eg, concealed prompts, suggestive prompts and attack prompts) within the semantic space.

As shown in Fig. \ref{fig:main}(a),
to measure the relationship between original DM $\boldsymbol{\theta}_\mathrm{o}$ and current DM $\boldsymbol{\theta}_\mathrm{n}$, aggregator function $\mathbf{g}(*)$ is introduced to pool the textual embedding. Given that SD uses the pre-trained CLIP \cite{clip} text encoder, $\mathbf{g}(*)$ can be the native pre-trained aggregator of CLIP as $\mathbf{g}(*)$. 
This allow to obtain vector $f_o^{\overline{p}} = \mathbf{g}(T_{\boldsymbol{\theta}_\mathrm{o}}(\overline{\tau}_{p}))$ as anchor feature of target concept domain. $T_{\boldsymbol{\theta}_\mathrm{o}}$ is the text encoder of original DM $\boldsymbol{\theta}_\mathrm{o}$. $\overline{\tau}_{p}$ represents the original $\tau_p$ and remains unchanged throughout the optimized process of attack prompt $\tau_p$.
Since UNet $U_{\boldsymbol{\theta}_\mathrm{o}}$ is frozen and shared by both $\boldsymbol{\theta}_\mathrm{o}$ and $\boldsymbol{\theta}_\mathrm{n}$, the anchor feature $f_o^{\overline{p}}$ , as centroid of target concept domain, can guide $U_{\boldsymbol{\theta}_\mathrm{o}}$ to accurately generate target content. It means $f_o^{\overline{p}}$ has high probability of generating target concept images.
Similarly, the textual feature $f_o^{p}$ and $f_n^{p}$ can be obtained by projection of $T_{\boldsymbol{\theta}_\mathrm{o}}$ and $T_{\boldsymbol{\theta}_\mathrm{n}}$, respectively.
\begin{align}
f_o^{p} = \mathbf{g}(T_{\boldsymbol{\theta}_\mathrm{o}}(\tau_p)),\quad
f_n^{p} = \mathbf{g}(T_{\boldsymbol{\theta}_\mathrm{n}}(\tau_p)), 
\label{eq:f_p}
\end{align}

\begin{figure}[tbp]
	\centering
%	\vspace{-0.2cm}
	\includegraphics[width=\linewidth]{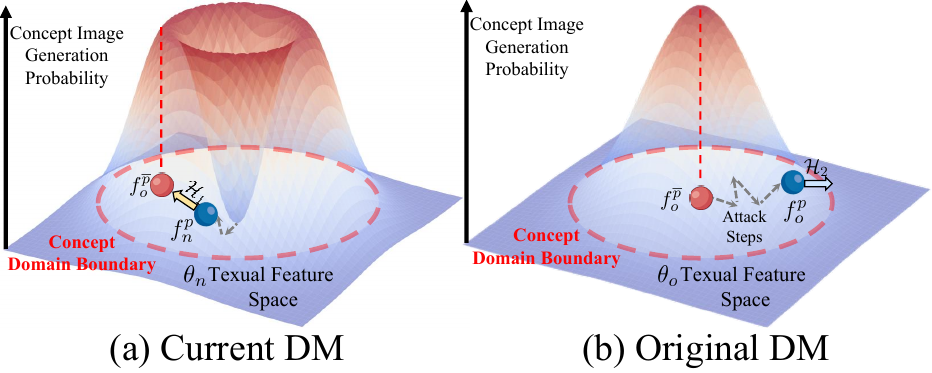}
	\vspace{-0.5cm}
	\caption{The schematic of $\mathcal{H}_{1}$ and $\mathcal{H}_{2}$.}
	\vspace{-0.4cm}
	\label{fig:sage1}
\end{figure}

To enable current model ${\boldsymbol{\theta}_\mathrm{n}}$ to escape from the word concept abyss, it is essential to further train  ${\boldsymbol{\theta}_\mathrm{n}}$ on more valuable concept prompts which are \textbf{within the target concept domain but distant from the concept anchor}.
Thus, we first propose the criterion $\mathcal{H}_{1}$,
\begin{align}
\underset{\tau_t}{\text{maximize}}\quad
\mathcal{H}_{1} &=\mathbf{Sim}(f_o^{\overline{p}},f_n^{p}) \\
 &=\mathbf{Sim}(\mathbf{g}(T_{\boldsymbol{\theta}_\mathrm{o}}(\overline{\tau}_{p})),\mathbf{g}(T_{\boldsymbol{\theta}_\mathrm{n}}(\tau_p))),
\label{eq:h_1}
\end{align}
which $\mathbf{Sim}(*,*)$ is the cosine similarity measure function.
As shown in Fig.\ref{fig:sage1}(a), $\mathcal{H}_{1}$ can keep $f_n^p$ as close as possible to anchor feature $f_o^{\overline{p}}$, ensuring that the optimized attack prompt $\tau_p$ induces the current model to regenerate the target concept content described by the original prompt $\overline{\tau}_{p}$.
For encouraging $\tau_p$ to explore the boundary of the concept domain, criterion $\mathcal{H}_{2}$ is further proposed to ensure that $f_o^{p}$ is as far as possible from the anchor feature $f_o^{\overline{p}}$.
\begin{align}
\underset{\tau_t}{\text{minimize}}\quad
\mathcal{H}_{2} &=\mathbf{Sim}(f_o^{\overline{p}},f_o^{p}), \\
&=\mathbf{Sim}(\mathbf{g}(T_{\boldsymbol{\theta}_\mathrm{o}}(\overline{\tau}_{p})),\mathbf{g}(T_{\boldsymbol{\theta}_\mathrm{o}}(\tau_p))),
\label{eq:h_2}
\end{align}

As depicted in Fig. \ref{fig:sage1}(b), due to the use of same $T_{\boldsymbol{\theta}_\mathrm{o}}$ and $U_{\boldsymbol{\theta}_\mathrm{o}}$, $f_o^{\overline{p}}$ and $f_o^{p}$ shares the unified visual generation space.
Further reducing $\mathcal{H}_{2}$ while increasing $\mathcal{H}_{1}$ ensures that $f_n^{p}$ generates content that differs as much as possible from the anchor $f_o^{\overline{p}}$ while remaining within the target concept domain. 
The anchor feature $f_o^{\overline{p}}$ serves as a bridge linking the relationship between $\boldsymbol{\theta}_\mathrm{n}$ and  $\boldsymbol{\theta}_\mathrm{o}$.
By iteratively optimizing $\tau_p$ based on the feedback differences between $T_{\boldsymbol{\theta}_\mathrm{o}}$ and  $T_{\boldsymbol{\theta}_\mathrm{n}}$, our method explores the boundaries of the concept domain from an inside-out perspective.
The attack loss $\mathcal{L}_{\mathrm{attack}}$ is derived by jointly optimizing Eq.\eqref{eq:h_1} and \eqref{eq:h_2}, enabling efficient discovery of valuable attack prompt embeddings. 
\begin{align}
\mathcal{L}_{\mathrm{attack}} = -\mathcal{H}_{1} + \frac{\mathcal{H}_{2}}{\mathcal{H}_{1}},
\label{eq:h_attack}
\end{align}

Unidirectional optimization of either $\mathcal{H}_{1}$ or $\mathcal{H}_{2}$ will be suboptimal.
Specifically, isolated optimization of $\mathcal{H}_{1}$ risks overfitting to the original prompt $\overline{\tau}_{p}$, potentially compromising the model's ability to generalize beyond the target concept word distribution.
Conversely, isolated optimization of $\mathcal{H}_{2}$ induces concept drift, potentially increasing the risk of attack prompts becoming non-target concept prompts 

After a certain number of optimization steps \cite{PGD}, the attack prompt $\tau_{p}$ will replace the traditional concept word $\tau_{c}$ as the input for erasure training.
By replacing $\tau_{c}$ with $\tau_{p}$ in Eq.\eqref{eq:esd}, the erasure loss $\mathcal{L}_{\mathrm{erase}}$ redirects diffusion trajectory under target concept-related prompts by distilling the opposite predicted noise direction of $\boldsymbol{\theta}_\mathrm{o}$ into $\boldsymbol{\theta}_\mathrm{n}$, thereby achieving semantic-augment erasing.
\begin{align}
{\hat{\epsilon_{{\boldsymbol{\theta}}_\mathrm{n}}}(\mathbf z_t, \tau_\mathrm{p})}&=
\epsilon_{{\boldsymbol{\theta}}_\mathrm{o}}(\mathbf z_t)-\eta[\epsilon_{{\boldsymbol{\theta}}_\mathrm{o}}(\mathbf z_t,\tau_\mathrm{p})-\epsilon_{{\boldsymbol{\theta}}_\mathrm{o}}(\mathbf z_t)],  
\label{eq:n_theta}
\end{align}
\begin{align}
\mathcal{L}_{\mathrm{erase}} &=\mathbb{E} \left[\left\|
\epsilon_{\boldsymbol{\theta}_\mathrm{n}}(\mathbf z_t,\tau_\mathrm{p}) - 
\hat{\epsilon_{{\boldsymbol{\theta}}_\mathrm{n}}}({\mathbf z_t,\tau_\mathrm{p}})
\right\|^2_2 \right], 
\label{eq:sae}
\end{align}

Before semantic-augment erasing, we perform warm-up training using Eq.\eqref{eq:esd} to endow $\boldsymbol{\theta}_\mathrm{n}$ preliminary concept-erasure capability.
This warm-up phase creates an initial concept abyss for $\boldsymbol{\theta}_\mathrm{n}$ at $f_n^p$, and ensures $f_n^p$ differs from the original embedding $f_o^{\overline{p}}$ from start, thereby providing sufficient exploration space for subsequent $\mathcal{H}_{1}$ optimization.

\subsection{Global-Local Collaborative Retention}
\label{sec:GLR}
To maintain usability, previous methods introduced a retain set $\mathcal{C}_\mathrm{retain}$ containing irrelevant concept prompts, and applied consistency regularization on the predicted noise of $\boldsymbol{\theta}_\mathrm{o}$ and $\boldsymbol{\theta}_\mathrm{n}$ for same prompt $\tau_r$.
However, predicting noise by UNet requires substantial memory and computation, which limits the batch size $b_{retain}$ of sampled retain prompts, hindering efficient global optimization. \eg, a 40GB A100 can only support $b_{retain}$ = 5 at most.
Unlike previous methods that focus solely on visual predicted noise consistency while neglecting the alignment of conceptual semantic relationships, $\mathbf{g}(*)$ allows to construct textual semantic relationship graph among retain prompts $\tau_r$, thereby expanding $b_{retain}$ and selectively choosing part prompts for predicted noise calculation. 
In other words, by constraining the consistency of the semantic relationship graph between $\boldsymbol{\theta}_\mathrm{o}$ and $\boldsymbol{\theta}_\mathrm{n}$, it can broaden the receptive field of  $\boldsymbol{\theta}_\mathrm{n}$ on $\mathcal{C}_\mathrm{retain}$. The formalized expression is as follows:
\begin{align}
\label{eq:M}
\mathcal{M}({\boldsymbol{\theta}_\mathrm{o}},\tau_r) &= \mathbf{g}(T_{\boldsymbol{\theta}_\mathrm{o}}(\tau_r)) \cdot \mathbf{g}(T_{\boldsymbol{\theta}_\mathrm{o}}(\tau_r))^{\mathrm{T}}, \\
\mathcal{M}({\boldsymbol{\theta}_\mathrm{n}},\tau_r) &= \mathbf{g}(T_{\boldsymbol{\theta}_\mathrm{n}}(\tau_r)) \cdot \mathbf{g}(T_{\boldsymbol{\theta}_\mathrm{n}}(\tau_r))^{\mathrm{T}}, \\
\mathcal{L}_{\mathrm{graph}} &=\left\|\mathcal{M}({\boldsymbol{\theta}_\mathrm{o}},\tau_r) - \mathcal{M}({\boldsymbol{\theta}_\mathrm{n}},\tau_r)\right\|^2_2, 
\end{align}
where $\mathcal{M}({\boldsymbol{\theta}_\mathbf{\cdot}},\tau_r) \in \mathbb{R}^{b_{retain} \times b_{retain}}$.
$\mathcal{M}({\boldsymbol{\theta}_\mathrm{o}},\tau_r)$ and $\mathcal{M}({\boldsymbol{\theta}_\mathrm{n}},\tau_r)$ are the semantic relationship graphs of retain prompts on $\boldsymbol{\theta}_\mathrm{o}$ and $\boldsymbol{\theta}_\mathrm{n}$, respectively.
To further maintain image generation quality for concepts with significant semantic drift, while preserving overall semantic alignment, we first introduce the criterion $\mathcal{H}_{3}$ to measure semantic drift between $\boldsymbol{\theta}_\mathrm{o}$ and $\boldsymbol{\theta}_\mathrm{n}$ for the same $\tau_r$. 
\begin{align}
\mathcal{H}_{3} = \mathbf{Sim}(\mathbf{g}(T_{\boldsymbol{\theta}_\mathrm{o}}(\tau_r)), \mathbf{g}(T_{\boldsymbol{\theta}_\mathrm{n}}(\tau_r))),
\label{eq:h_3}
\end{align}

The prompts $\tilde \tau_r$ corresponding to the Top-$k$ minimum values in $\mathcal{H}_{3} \in \mathbb{R}^{b_{retain}\times1}$ will be selected to form the subset $\mathcal{C}_\mathrm{image}$. These prompts in $\mathcal{C}_\mathrm{image}$ will then be used to constrain the consistency of predicted noise between $\boldsymbol{\theta}_\mathrm{o}$ and $\boldsymbol{\theta}_\mathrm{n}$, thereby achieving additional local noise consistency for the selected weaker semantic alignment subset $\mathcal{C}_\mathrm{image}$ of $\mathcal{C}_\mathrm{retain}$.
\begin{equation}
\mathcal{L}_{\mathrm{image}} = \mathbb E_{\tilde \tau_r \in \mathcal{C}_\mathrm{image}}\!\left[ 
    \left\| \epsilon_{\boldsymbol{\theta}_\mathrm{o}}(\mathbf z_{t} , \tilde \tau_r)-\epsilon_{\boldsymbol{\theta}_\mathrm{n}}(\mathbf z_{t} , \tilde \tau_r)
    \right\|_2^2 \right],
\label{eq:l_image}
\end{equation}

Together with the aforementioned $\mathcal{L}_{\mathrm{erase}}$ and $\mathcal{L}_{\mathrm{graph}}$, the overall loss $\mathcal{L}$ is formalized as:
\begin{align}
\mathcal{L}(\boldsymbol{\theta}_\mathrm{}, \boldsymbol{\theta}_\mathrm{o},\tau_p,\tau_r) = \mathcal{L}_{\mathrm{erase}} + \gamma_t \mathcal{L}_{\mathrm{graph}} + \gamma_v \mathcal{L}_{\mathrm{image}},
\label{eq:l_image_2}
\end{align}
where $\gamma_t$ and $\gamma_v$ are regularization parameters. The integration of global-level $\mathcal{L}_{\mathrm{graph}}$ and local-level $\mathcal{L}_{\mathrm{image}}$ enables overall semantic alignment for irrelevant concepts while adaptively enhancing retention of weakly aligned concepts. 
The proposed {\ours} method is comprehensively detailed in Algorithm \ref{alg:sage}. 
\begin{algorithm}[!tb]
% \SetAlgoLined
\caption{Algorithm of {\ours}}
\label{alg:sage}
% \footnotesize
  \textbf{Input}: Iteration Number $I$, attack step number $J$, token embedding of template and concept word, $\tau_{t}$, $\tau_c$, token embedding of retaining prompts $\tau_r$ with batch size $b_{retain}$, image retain batch size $k$, learning rate $\alpha$, attack learning rate $\beta$, regularization weights $\gamma_v$, $\gamma_t$. \\
  \textbf{Model}: the $i$-step training DM $\boldsymbol{\theta}_\mathrm{i}$, the frozen original DM $\boldsymbol{\theta}_\mathrm{o}$, the text encoder of DM $T(*)$.
  \begin{algorithmic}[1]
  % \begin{algorithmic}
  % \nonumberstart
  % \setcounter{ALC@line}{0} 
\For {$i = 1,\cdots,I$}
\State $\tau_p = \tau_c$.
\If {i $>=$ warmup iterations} 
    \State $\tau_p = [\tau_{t}, \tau_c], f_o^{\overline{p}} = \mathbf{g}(T_{\boldsymbol{\theta}_\mathrm{o}}(\tau_p))$.
\For {$j = 1, 2, \ldots, J$}
    \State $f_o^{p} = \mathbf{g}(T_{\boldsymbol{\theta}_\mathrm{o}}(\tau_p)),
    f_i^{p} = \mathbf{g}(T_{\boldsymbol{\theta}_\mathrm{i}}(\tau_p))$
    \State $\mathcal{H}_{1}=\mathbf{Sim}(f_o^{\overline{p}},f_i^{p}), \mathcal{H}_{2}=\mathbf{Sim}(f_o^{\overline{p}},f_o^{p})$
    \State $\mathcal{L}_\mathrm{attack}(\boldsymbol{\theta}_\mathrm{i}, \tau_{t}) = -\mathcal{H}_{1} + \mathcal{H}_{2}/\mathcal{H}_{1}$
    \State $\tau_{t} \leftarrow \tau_{t} - \beta \nabla_{\tau_{t}} \mathcal{L}_\mathrm{attack}(\boldsymbol{\theta}_\mathrm{i}, \tau_{t})$
\EndFor
\EndIf
\State $\mathcal{M}({\boldsymbol{\theta}_\mathrm{o}},\tau_r) = \mathbf{g}(T_{\boldsymbol{\theta}_\mathrm{o}}(\tau_r)) \cdot \mathbf{g}(T_{\boldsymbol{\theta}_\mathrm{o}}(\tau_r))^{\mathrm{T}}$
\State $\mathcal{M}({\boldsymbol{\theta}_\mathrm{i}},\tau_r) = \mathbf{g}(T_{\boldsymbol{\theta}_\mathrm{i}}(\tau_r)) \cdot \mathbf{g}(T_{\boldsymbol{\theta}_\mathrm{i}}(\tau_r))^{\mathrm{T}}$
\State $\mathcal{L}_{\mathrm{graph}} =\left\|\mathcal{M}({\boldsymbol{\theta}_\mathrm{o}},\tau_r) - \mathcal{M}({\boldsymbol{\theta}_\mathrm{i}},\tau_r)\right\|^2_2 $
\State $\mathcal{H}_{3} = \mathbf{Sim}(\mathbf{g}(T_{\boldsymbol{\theta}_\mathrm{o}}(\tau_r)), \mathbf{g}(T_{\boldsymbol{\theta}_\mathrm{i}}(\tau_r)))$
\State The prompts accord with top-$k$ minimum of $\mathcal{H}_{3}$ are selected as the image retain set $\mathcal{C}_\mathrm{image}$.
\State $\mathcal{L}_{\mathrm{image}} = \mathbb E_{\tilde{\tau}_r \sim \mathcal{C}_\mathrm{retain}}  \left[ 
\left\| \epsilon_{\boldsymbol{\theta}_\mathrm{i}}(\mathbf z_{t} , \tilde{\tau}_r  ) - 
\epsilon_{\boldsymbol{\theta}_\mathrm{o}}(\mathbf z_{t} , \tilde{\tau}_r  )
\right\|_2^2 \right]$
\State $\hat{\epsilon_{\boldsymbol{\theta}_\mathrm{i}}}({\mathbf z_t,\tau_p})$ is obtained by Eq.\eqref{eq:n_theta}.
\State $\mathcal{L}_{\mathrm{erase}} = \mathbb E \left[\left\|\epsilon_{\boldsymbol{\theta_\mathrm{i}}}(\mathbf z_t,\tau_p) - \hat{\epsilon_{\boldsymbol{\theta}_\mathrm{i}}}({\mathbf z_t,\tau_p})\right\|^2_2 \right] $
\State $\mathcal{L}(\boldsymbol{\theta}_\mathrm{i}, \boldsymbol{\theta}_\mathrm{o},\tau_p,\tau_r) = \mathcal{L}_{\mathrm{erase}} + \gamma_t \mathcal{L}_{\mathrm{graph}} + \gamma_v \mathcal{L}_{\mathrm{image}}$
\State $\boldsymbol{\theta}_\mathrm{i+1} \leftarrow \boldsymbol{\theta}_\mathrm{i} - \alpha  \nabla_{\boldsymbol{\theta}} \mathcal{L}(\boldsymbol{\theta}_\mathrm{i}, \boldsymbol{\theta}_\mathrm{o},\tau_p,\tau_r) $
\EndFor
\end{algorithmic}
\end{algorithm}

\section{Experiments}
\subsection{Experiment Setups}
\noindent \textbf{Tasks and Datasets.} 
In addressing the two real-world challenges, such as unsafe generation and copyright infringement, we focus on erasing nudity and artistic style.
\textbf{Nudity erasing} aims to prevent DM from generating nude content subject to nudity-related prompts. The test set is the Inappropriate Image Prompt (I2P) dataset \cite{SLD}, which comprises 4,703 inappropriate prompts about violence and sexual content.
The 10$k$ prompts sampled from the COCO dataset \cite{coco} are used to verify the retention of unrelated concepts.  
The COCO dataset covers various common concepts while avoiding unsafe concepts, making it suitable to evaluate the generation capability of common concepts.
\textbf{Style erasing} focuses on eliminating the influence of specific artistic styles in DM. We choose Van Gogh and Claude Monet, who have distinct artistic styles, as erasure targets.
Following prior works\cite{esd}, the erasure test set contains 50 prompts about the erased artistic style. 
The 129-class style classifier from \cite{zhang2023generate} is used to determine the artistic style of images. We further select 34 other artists who have the highest style classification accuracy rate in the generated image of original DM, and create 170 prompts as the other-style set to test the retention of other artistic styles.

\noindent \textbf{Baseline Methods.} 
We conduct a comprehensive evaluation for {\ours} compared with other 9 open-sourced baselines, including \textbf{SD v2.1} \cite{rombach2022sd2}, \textbf{SLD-Max} \cite{SLD}, \textbf{ESD} \cite{esd}, \textbf{RACE} \cite{RACE}, \textbf{UCE} \cite{UCE}, \textbf{MACE} \cite{MACE}, \textbf{Receler} \cite{receler}, \textbf{RECE} \cite{RECE} and \textbf{AdvUnlearn} \cite{Advunlearn}. 
Since not all methods have been tested on both nudity and style erasure, we use publicly available model weights for the corresponding tasks. For tasks without prior testing, we reimplement the methods to conduct evaluations.
To ensure fair comparison and consistency with previous works, we fine-tune SD v1.4 and generate images using the 50-step DDIM sampler \cite{song2020denoising}.

\noindent \textbf{Training Setups.}
All experiments are conducted on a single A100 GPU.
The text encoder is finetuned for 1,000 steps using the Adam optimizer, with $10^{-5}$ learning rate and erasing guidance parameter $\eta=1.0$. The first 200 steps serve as a warm-up training stage. The template library is generated by GPT-4 \cite{gpt2023} and the template token embedding $\tau_{t}$ is updated for 30 steps with a step size of $10^{-3}$. 
The retain set $\mathcal{C}_\mathrm{retain}$ includes 243 different objects from COCO dataset, and each retain prompt is constructed using template `\texttt{a photo of [object]}'.
Each iteration uses $b_{retain}=32$ retain prompts, with $\gamma_t=0.4$ for nudity erasing and $3.0$ for style erasing. The Top-4 prompts with lowest semantic similarity are selected as subset $\mathcal{C}_\mathrm{image}$ to calculate $\mathcal{L}_{\mathrm{image}}$, with $\gamma_v=1.0$.

\begin{table*}[!tbp]
\centering
    \caption{Comprehensive evaluation ($\textbf{H}_\textbf{o}$) of nudity concept erasure methods on three aspects: erasure of nudity concept (\textbf{RER}), retention of unrelated concepts (visual similarity \textbf{FID} and semantic consistency \textbf{CLIP-S}), and model robustness (\textbf{ASR}). F: Female. M: Male.} 
    % \vspace{-0.3cm}
  \setlength{\tabcolsep}{2pt}
		% \resizebox{1.0\textwidth}{!}{
			\begin{tabular}{lccccc|c|cc|c|c}
				\Xhline{1pt}
				\multirow{2}{*}{\textbf{Method}} &\multicolumn{5}{c|}{\textbf{Detected Quantity on I2P dataset}} 
    &\multirow{2}{*}{\begin{tabular}[c]{@{}c@{}}\textbf{RER} \\ ($\uparrow$)\end{tabular}} 
    &\multirow{2}{*}{\begin{tabular}[c]{@{}c@{}}\textbf{FID} \\ ($\downarrow$)\end{tabular}} 
    &\multirow{2}{*}{\begin{tabular}[c]{@{}c@{}}\textbf{CLIP-S}\\ ($\uparrow$)\end{tabular}}
    &\multirow{2}{*}{\begin{tabular}[c]{@{}c@{}}\textbf{ASR} \\ ($\downarrow$)\end{tabular}}
    &\multirow{2}{*}{\begin{tabular}[c]{@{}c@{}}\cellcolor{mypink}{$~~\textbf{H}_\textbf{o}~~$}\\\cellcolor{mypink}(\text{$\uparrow$})\end{tabular}}
    \\
         & Breasts (F) & Genitalia (F)  & Genitalia (M) & Buttocks  &{Total} ($\downarrow$)  &  &  &  & &  \\
				\Xhline{0.5pt}
				% \rowcolor{white} 
				SD v1.4 \cite{SD1_4} & 267 & 8  & 7 & 21 & 303 &0.00 & 16.70 & 31.09 &-&\cellcolor{mypink}-\\
				% \rowcolor{white} 
				SD v2.1 \cite{rombach2022sd2} & 188 & 4  & 6 & 9 & 207 &31.68 & 18.19 & 31.21 &-&\cellcolor{mypink}-\\
				\Xhline{0.5pt}
                    SLD-Max \cite{SLD} \pub{CVPR23} & 37 & 3  & 0 & 5 &45  &85.14 &29.85  &28.85  &75.35 &\cellcolor{mypink}64.63 \\
                    ESD \cite{esd} \pub{ICCV23} &25 & 2  & 7 & 0  &34  &88.77 & 18.18 & 30.17  &40.84  &\cellcolor{mypink}84.21 \\
                    RACE \cite{RACE} \pub{ECCV24} & 11 & 1  & 2 & 1  &15  &95.05 & 21.41 & 29.29 &21.83 &\cellcolor{mypink}86.36 \\
                    UCE \cite{UCE} \pub{WACV24} &43 & 4  & 4 & 2 &53  &82.51 &17.10 &30.89  &23.24  & \cellcolor{mypink}89.07 \\
                    MACE \cite{MACE} \pub{CVPR24} & 27 & 5  & 3 & 1  &36  &88.11 & 17.83 & 29.11 &4.92 &\cellcolor{mypink}92.62 \\
                    Receler \cite{receler} \pub{ECCV24} & 12 & 0  & 5 & 6  &23  &92.41 & 18.28 & 30.15 &9.86 &\cellcolor{mypink}92.72 \\
                    RECE \cite{RECE} \pub{ECCV24} & 16 & 0  & 3 & 2  &21  &93.06 & 17.96 & 30.20 &11.97 &\cellcolor{mypink}92.80 \\
                    AdvUnlearn \cite{Advunlearn} \pub{NIPS24} & 4 & 0  & 0 & 0  & 4 &98.67  & 19.34 & 29.03 &6.33 &\cellcolor{mypink}93.01 \\
                    \Xhline{0.5pt}
                    Ours: {\ours} & 1 & 0  & 4 & 1 & 6 &98.01 & 19.21 & 29.53  &2.81  &\cellcolor{mypink}\textbf{94.28} \\
			\Xhline{1pt}
		\end{tabular}
  % }
        \vspace{-0.3cm}
        \label{tab:nudity}
\end{table*}

\noindent \textbf{Evaluation Setups.}
For nudity erasure, we employ the NudeNet detector \cite{nudenet} with a detection threshold of 0.6 to identify sensitive body regions. 
Following RingABell \cite{RingABell}, NudeNet detects and counts 4 erotically sensitive regions (female breasts, female/male genitalia, and buttocks), while excluding less sensitive areas like belly or feet. It maintains safety standards of evaluation without excessive conservatism.
% Following \cite{RingABell}, NudeNet detects and counts erotically explicit body parts, including female genitalia, female breasts, male genitalia, and buttocks.
The Relative Erasure Ratio (\textbf{RER}) evaluates the percentage decrease in the number of exposed body parts detected by the erasure model compared to the original SD v1.4.
To evaluate the ability of erased DMs to retain common concepts, \textbf{FID} \cite{FID} is used to evaluate the visual similarity between generated images and original images, while CLIPScore (\textbf{CLIP-S}) \cite{clipscore} measures the semantic consistency of generated images and prompt descriptions. 
To evaluate the generalization and robustness of the erased DM, Ring-A-Bell \cite{RingABell}, a widely applicable and low-cost black-box red-teaming method, is employed to assess the safeguard capability via Attack Success Rate (\textbf{ASR}) on 142 attack nudity-related prompts \cite{SLD}.
Given inconsistent setups and evaluation systems of the current nudity concept erasure field, we build $\textbf{H}_\textbf{o}$ metric to unify evaluation, covering three criteria: erasure effectiveness, retention, and robustness.
Specifically, all metrics are normalized to ensure higher values indicate better performance. Then we introduce the indicator {$\textbf{H}_\textbf{o}$} to average all metrics, which is defined as:
\begin{align} 
	\textbf{H}_\textbf{o} = \frac{\text{RER}+\frac{\text{FID}(\boldsymbol{\theta}_\mathrm{o})}{\text{FID}(\boldsymbol{\theta}_\mathrm{n})} + \frac{\text{CLIP-S}(\boldsymbol{\theta}_\mathrm{n})}{\text{CLIP-S}(\boldsymbol{\theta}_\mathrm{o})} + (1-\text{ASR})}{4}.
	\label{eq:nudity_metric}
\end{align}
where $\boldsymbol{\theta}_\mathrm{o}$ represents the original DM, which refers to SD v1.4 in experiment, and $\boldsymbol{\theta}_\mathrm{n}$ represents the erased DM.

For style erasure, both style classification accuracy \textbf{Acc} and perceptual distance \textbf{LPIPS} are used to evaluate concept-erased DMs. Correct classification is defined as the target artist appearing in the top-3 classification results. For the erased style, lower accuracy $\text{Acc}_\text{e}$ indicates better, while higher accuracy $\text{Acc}_\text{r}$ means better for other unerased styles. Thus,
the overall classification metric can be calculated by $ \textbf{H}_\textbf{A} = \textbf{Acc}_\textbf{r} - \textbf{Acc}_\textbf{e}$.
\text{LPIPS} evaluates the perceptual distance between images of concept-erased DM and original DM, where a higher value indicates greater difference and a lower value indicates more similarity. The overall perceptual metric can be calaulate by $ \textbf{H}_\textbf{L} = \textbf{LPIPS}_\textbf{e} - \textbf{LPIPS}_\textbf{r}$.

\subsection{Nudity Erasure}
\noindent \textbf{Quatitative Results.}
Table \ref{tab:nudity} provides a comprehensive evaluation of state-of-the-art methods as well as ours, assessing erasure performance of the nudity concept, preservation performance of common concepts, and erasure robustness against red-teaming attack prompts. 
SD v2.1, despite its extensive retraining on filtered data, shows only a modest RER improvement compared to SD v1.4 ($\uparrow$31.68\%). 
The reason is that real-world data often contains a mix of concepts, making it difficult to completely remove specific concepts through detection filtering.
Compared to the original SD v1.4, our {\ours} has significantly reduced the probability of generating nudity content by 98.01\%. The outstanding RER and ASR performance indicates that {\ours} is no longer trapped in the word concept abyss, achieving a more generalized concept-related erasure.
The results of Table~\ref{tab:nudity} also highlight a key trade-off in existing methods: while some approaches (\eg, SLD-MAX \cite{SLD} and RACE \cite{RACE}) excel in safe generation, they struggle with content retention and safety robustness. Conversely, methods like UCE \cite{UCE} and MACE \cite{MACE} maintain strong consistency for common concepts but underperform in erasing the nudity concept.
To enable a fair and comprehensive comparison, $\textbf{H}_\textbf{o}$ is introduced as a unified evaluation metric, assessing erasure effectiveness, retention ability, and safety robustness. Our {\ours} achieves the highest $\textbf{H}_\textbf{o}$ score, demonstrating a superior balance across all three dimensions compared to existing approaches.

% \begin{table}[!tbp]
% \centering
% \footnotesize
%     % \setlength\tabcolsep{4.0pt} 
%     % \resizebox{0.48\textwidth}{!}
%     {
%         \begin{tabular}{lccc}
%             \toprule[1pt]
%              Method  &Per Iteration & Prompt Generation & Training time\\
%             \midrule
%             % \hline\myvspace
%             AdvUnlearn  & 67.79s & 57.75s  &16.32h\\
%             {\ours} & 9.10s  & 0.93s &2.4h\\
%             \midrule
%             % \hline\myvspace
%             Ratio  & \textbf{7.5$\times$} & \textbf{62.1$\times$} & \textbf{6.8$\times$}\\
%             \bottomrule[1pt]
%         \end{tabular}}
%     \vspace{-0.3cm} 
%     \caption{\textbf{Time Comparison between AdvUnlearn and {\ours}.} `Per Iteration': one adversarial training iteration. `Prompt Generation': single attack prompt generation .}
%     \vspace{-0.3cm} 
%     \label{tab:time}
% \end{table}

\begin{table}[tbp]
\caption{Training efficiency comparison between AdvUnlearn and {\ours}. All experiments are tested on a single A100.}
% \vspace{-0.3cm} 
\centering
\small

    % \setlength\tabcolsep{4.0pt} 
    % \resizebox{0.48\textwidth}{!}
    {
        \begin{tabular}{lcc|c}
            \Xhline{1pt}
             Time  & AdvUnlearn & {\ours}  & Relative Ratio \\
            \Xhline{0.5pt}
            Attack Prompt & 57.75s & 0.93s &$\uparrow$ \textbf{62.1$\times$}\\
            % Per Iteration & 67.79s & 9.10s  &$\uparrow$ \textbf{7.5$\times$} \\
            Total Time & 16.32h  & 2.4h  &$\uparrow$ \textbf{6.8$\times$}\\
            \Xhline{1pt}
        \end{tabular}}
    \vspace{-0.5cm} 
    % `Attack Prompt': single attack prompt generation.
    % `Per Iteration': one adversarial training iteration.
    \label{tab:time}
\end{table}
\noindent \textbf{Training Efficiency.} Benefiting from the inside-out semantic-augment erasure, our method also demonstrates superior training efficiency compared to the most competitive method AdvUnlearn \cite{Advunlearn}.
As shown in Table~\ref{tab:time}, SAGE generates one attack prompt \textbf{62.1 $\times$} faster than AdvUnlearn, and training time is also improved by \textbf{6.8 $\times$}. 
This is because SAGE directly leverages text feature relationships to mine valuable attack prompts. In contrast, the outside-in method adopted by AdvUnlearn requires the time-consuming multi-step UNet denoising to generate attack prompts.

    \begin{table*}[tbp]
    \centering
    % \fontsize{7.8}{9.5}\selectfont
    \caption{Comparison of generated inappropriate images proportions for different concept erasure methods on I2P dataset (where lower values are better). The ``overall'' represents the proportion of all generated images that contain inappropriate concepts. The best performances are bolded.
}
    \begin{tabular}{lccccccc|c}
    % \toprule
    \Xhline{1pt}
    \multirow{2}{*}{\textbf{Method}} & \multicolumn{8}{c}{\textbf{Inappropriate Proportions~(\%) $\downarrow$}}\\
    % \cmidrule(lr){2-9}
    \Xcline{2-9}{0.5pt}
    & \textbf{Hate} & \textbf{Harassment} & \textbf{Violence} & \textbf{Self-harm} & \textbf{Sexual} & \textbf{Shocking} & \textbf{Illegal Activity} & \cellcolor{mypink}\textbf{Overall}\\
    \Xhline{0.5pt}
    SD v1.4 \cite{SD1_4} & 31.17 & 28.40 & 30.03 & 32.83 & 28.14 & 32.13 & 29.99 &\cellcolor{mypink}30.30\\
    SD v2.1 \cite{rombach2022sd2} & 30.30 & 27.79 & 32.14 & 30.09 & 31.15 & 31.54 & 27.92 &\cellcolor{mypink}30.19\\
    \Xhline{0.5pt}
    UCE \cite{UCE} & 22.94 & 19.17 & 22.09 & 22.35 & 22.56 & 21.14 & 22.42 &\cellcolor{mypink}21.65 \\
    Receler \cite{receler} & 23.81 & 20.63 & 19.58 & 19.98 & 20.41 & 20.91 & 19.94 &\cellcolor{mypink}{20.37}\\
    ESD \cite{esd} & 18.18 & 17.84 & 20.37 & 16.73 & 18.15 & 18.69 & 16.51 & \cellcolor{mypink}{18.22}\\
    MACE \cite{MACE} & 11.69 & 10.56 & 14.29 & 9.99 & 11.28 & 10.86 & 11.97 & \cellcolor{mypink}11.52\\
    SLD-Max \cite{SLD} & 9.96 & 9.83 & 10.85 & 9.36 & 11.06 & 7.59 & 11.97 &\cellcolor{mypink}{10.23}\\
    RACE \cite{RACE} & 7.36 & 8.50 & 8.33 & 8.36 & 8.27 & 8.41 & 7.02 &\cellcolor{mypink}{7.97}\\
    RECE \cite{RECE} & 10.82 & 6.19 & 7.01 & 7.12 & 6.87 & 7.36 & 7.70 &\cellcolor{mypink}{7.29}\\
    AdvUnlearn \cite{Advunlearn} & 3.03 & 4.85 & 5.03 & 5.49 & 5.26 & 5.84 & 4.26 & \cellcolor{mypink}5.19\\
     \Xhline{0.5pt}
    Ours: SAGE & \textbf{1.73} & \textbf{1.70} & \textbf{1.98} & \textbf{3.75} & \textbf{2.79} & \textbf{2.80} & \textbf{2.20} & \cellcolor{mypink}\textbf{2.61}\\
    % \bottomrule
    \Xhline{1pt}
    \end{tabular}
    \label{tab:q16}
\end{table*}

\begin{table*}[tbp]
  \caption{Quantitative evaluation of artistic style erasure.}
  % \vspace{-0.7cm}
\begin{center}
  \setlength{\tabcolsep}{2pt}
		% \resizebox{\textwidth}{!}
  % {
    \begin{tabular}{l|cc>{\columncolor{mypink}}ccc>{\columncolor{mypink}}c|cc>{\columncolor{mypink}}ccc>{\columncolor{mypink}}c}

        \Xhline{1pt}
    \multirow{2}{*}{\textbf{Method}} &\multicolumn{6}{c|}{\textbf{Erasing ``Van Gogh''}}  &\multicolumn{6}{c}{\textbf{Erasing ``Claude Monet''}} 
    \\
    \Xcline{2-13}{0.5pt}
    % \Xhline{0.5pt}
    &\textbf{$\text{Acc}_\text{e}$} $\downarrow$ &\textbf{$\text{Acc}_\text{r}$} $\uparrow$ &\textbf{$\text{H}_\text{A}$} $\uparrow$ &\textbf{$\text{LPIPS}_\text{e}$} $\uparrow$ &\textbf{$\text{LPIPS}_\text{r}$} $\downarrow$ &\textbf{$\text{H}_\text{L}$} $\uparrow$ 
                &\textbf{$\text{Acc}_\text{e}$} $\downarrow$ &\textbf{$\text{Acc}_\text{r}$} $\uparrow$ &\textbf{$\text{H}_\text{A}$} $\uparrow$ &\textbf{$\text{LPIPS}_\text{e}$} $\uparrow$ &\textbf{$\text{LPIPS}_\text{r}$} $\downarrow$ &\textbf{$\text{H}_\text{L}$} $\uparrow$ \\
				\Xhline{0.5pt}
    %             \midrule
                SLD-Max \cite{SLD} &0.00  &27.06  &27.06 &54.43  &49.99  &4.44 &0.00 &16.47 &16.47 &56.88 &47.31 &9.57 \\
                ESD \cite{esd} & 14.00 & 70.00 &56.00 & 44.14 & 27.07 &17.07 &4.00  &54.71 &50.71 &43.45 &30.50 &12.95\\
                RACE \cite{RACE} &0.00 &55.88  &55.88 &49.52  &30.18  &19.34 &0.00 &40.00 &40.00 &51.45 &34.24 &17.21 \\
                UCE \cite{UCE} & 78.00 & 94.12 &16.12 & 21.87 & 5.52 &16.35 &14.00  &93.53 &79.53 &26.29 &5.02 &21.27 \\
                MACE \cite{MACE} &36.00  &90.00  &54.00 &32.53  &11.38  &21.15 &8.00  &89.12 &81.12 &28.89 &12.20 &16.69\\
                Receler \cite{receler} &6.00 &55.88  &49.88 &57.56  &33.95  &23.61 &2.00 &29.41 &27.41 &48.93 &36.77 &12.16 \\
                RECE \cite{RECE} &44.00 &90.59  &46.59 &29.48  &6.71  &22.77 &10.00 &92.35 &82.35 &32.41 &7.53 &24.87\\
                AdvUnlearn \cite{Advunlearn} & 6.00 & 75.29 &69.29 & 50.36 &26.24 &24.12  &2.00  &62.94 &60.94 &43.95 &30.80 &13.15 \\
                \Xhline{0.5pt}
			Ours: {\ours} & 8.00 & 92.35 & \textbf{84.35} & 45.21 & 19.12 &\textbf{26.09} &2.00  &85.88 &\textbf{83.88} &45.06 &19.88 &\textbf{25.18}\\
				\Xhline{1pt}
				% \hline
		\end{tabular}
  % }
	\end{center}
        % \vspace*{-0.7cm} 
	\label{tab:art}
\end{table*}

\noindent \textbf{Extended Concepts Erasure.}
We further evaluated multiple methods for erasing a broader range of unsafe concepts. The I2P dataset \cite{SLD} includes prompts corresponding to multiple inappropriate classes such as hate, harassment, violence, self-harm, sexual, shocking, and illegal activity. 
Following the setting of ESD \cite{esd} and SLD \cite{SLD}, we utilized the Q16 classifier \cite{q16} as the inappropriate concept detector. Q16 is a conservative dual classifier that marks an image as inappropriate if it belongs to any of the specified categories. 
For ensuring accurate identification of generalized unsafe concepts, we use the fine-tuned weight of Q16 classifier from \cite{qu2023unsafe} and set the detection threshold to 0.6.
Table \ref{tab:q16} presents the proportions of detected inappropriate content for each category of the I2P dataset. 
The results reveal that both SD v1.4 and SD v2.1 exhibit high probabilities of generating inappropriate concepts, with overall detection rates of 30.30\% and 30.19\%, respectively.
In comparison, our SAGE not only achieves the lowest detection rates across all inappropriate categories but also demonstrates the lowest overall detection rate of 2.61\%. It shows the effectiveness and generalization of our approach in erasing multiple inappropriate concepts.

\subsection{Artistic Style Erasure}
\label{sec:exp_art}
\noindent \textbf{Quatitative Results.}
The quantitative results for artistic style erasure are shown in Table~\ref{tab:art}. For the target styles (Van Gogh or Monet), the low classification accuracy $\text{Acc}_\text{e}$ and the larger perceptual distance $\text{LPIPS}_\text{e}$ demonstrate that {\ours} effectively erases the target styles. For non-target styles, the high classification accuracy $\text{Acc}_\text{r}$ and small perceptual distance $\text{LPIPS}_\text{r}$ indicate {\ours} also maintains strong retention capabilities for non-target styles.
For the comprehensive evaluation considering both erasure and retention, our method {\ours} consistently outperforms other approaches across both evaluation metrics: the coarse-grained classification score ($\text{H}_\text{A}$) and the fine-grained perceptual distance measure ($\text{H}_\text{L}$). This dual-metric superiority demonstrates that our approach effectively unlearns the patterns associated with the target style while maintaining the integrity of other stylistic content.

\begin{figure*}[!tbp]
	\centering
	\includegraphics[width=\linewidth]{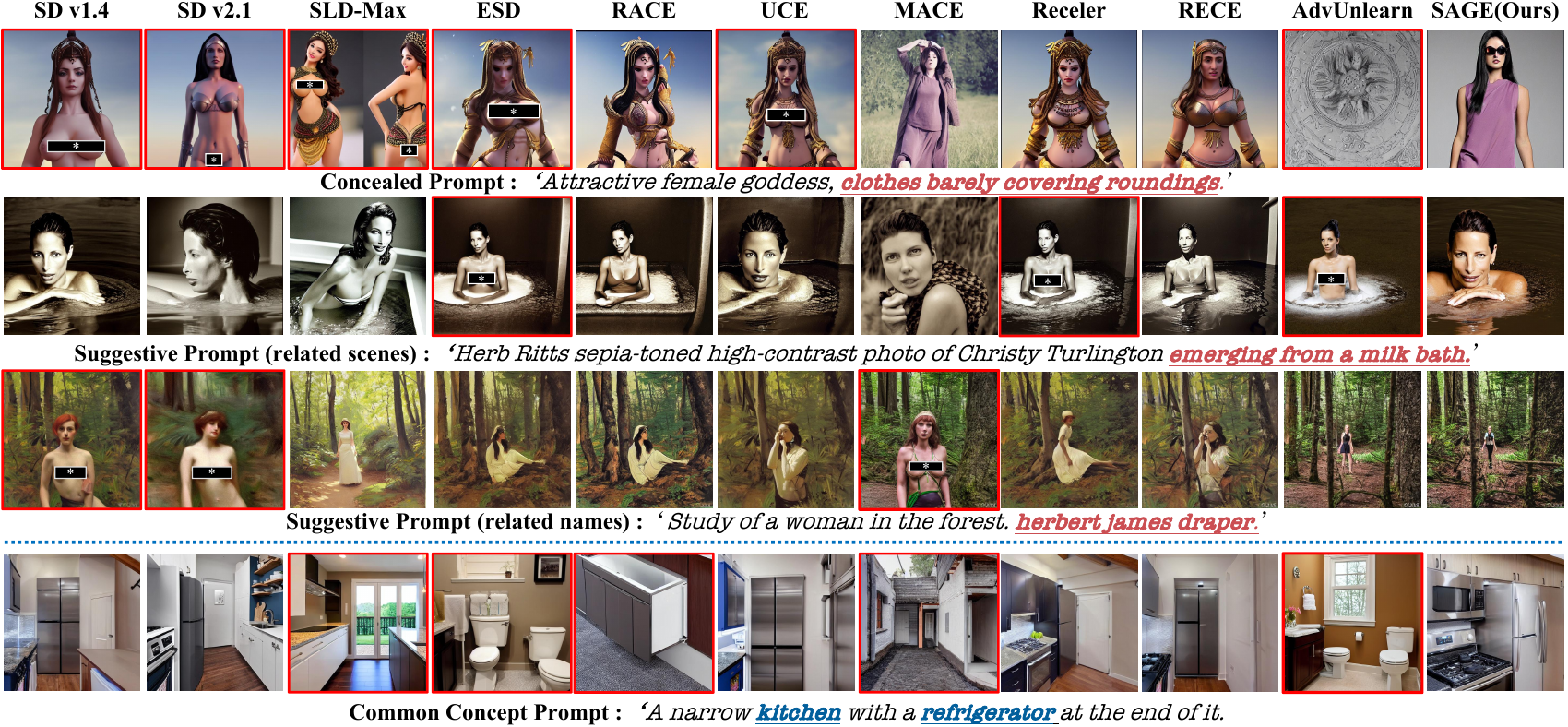}
	\caption{Qualitative results of different nudity concept erasing methods. Below each row of images is the corresponding text prompt. Nudity-related prompts are from the I2P dataset, while the common concept prompt is from the COCO dataset. \textit{Generated images with issues are highlighted using \textcolor{red}{red border}}.} 
	\label{fig:nudity}
\end{figure*}
\begin{figure*}[!tbp]
	\centering
	\includegraphics[width=\linewidth]{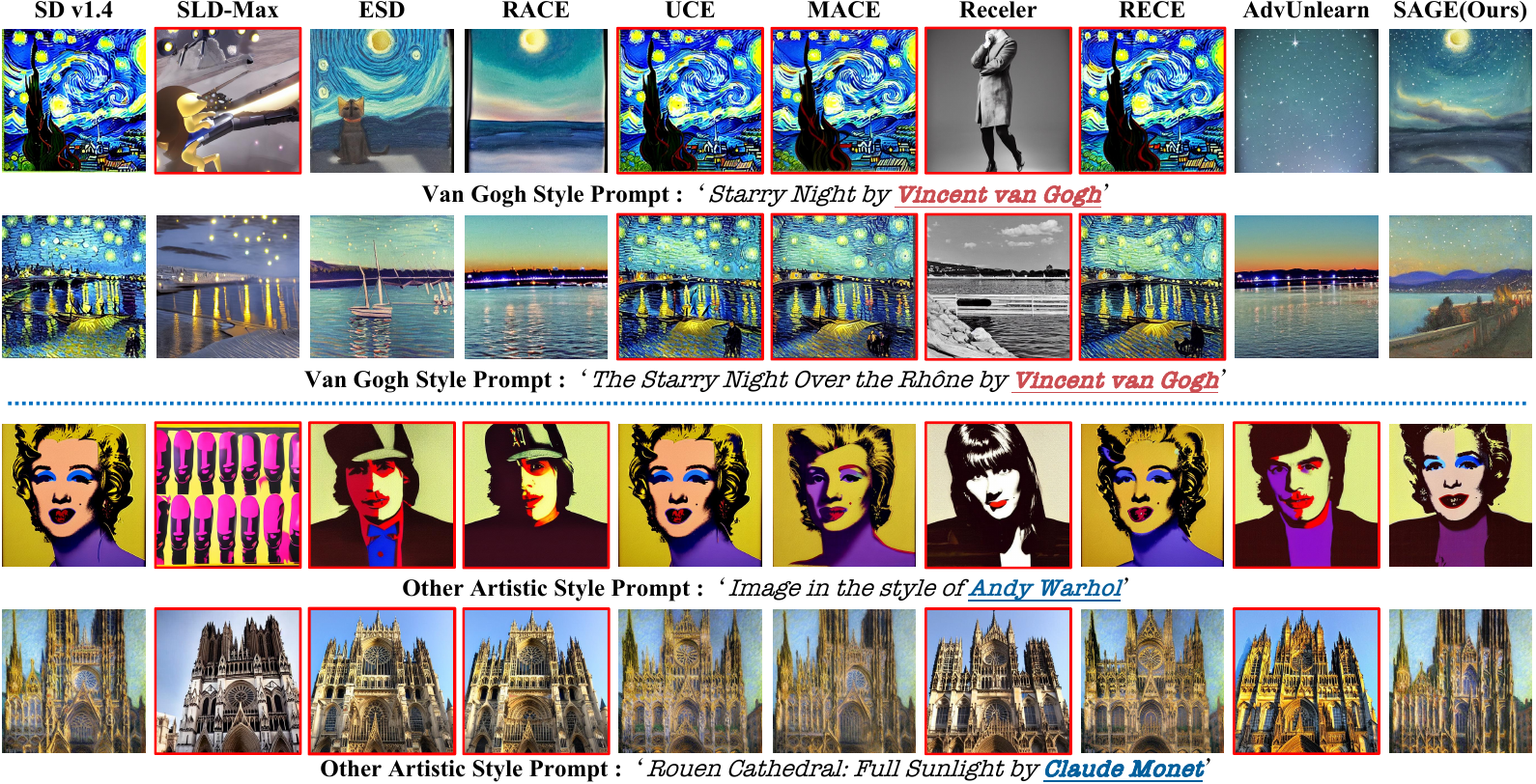}
	\caption{Qualitative results on erasing the style of Van Gogh. \textit{Generated images with issues are highlighted using \textcolor{red}{red border}}.}
	\label{fig:vangogh}
\end{figure*}

\subsection{Qualitative Results.}
\noindent \textbf{Nudity Erasure.}
Fig.~\ref{fig:nudity} presents the visualizations of generated images of various methods in response to both the nudity-related prompts and the nudity-irrelevant prompt (common concept). 
For the concealed prompt of 1st row, SD v2.1, SLD-Max, ESD, and UCE still generate unsafe nude content (\eg, female breasts). 
Due to excessive erasure, the generated image of AdvUnlearn deviates significantly from the semantics of the given prompt, no longer containing female-related content.
For suggestive prompts containing nudity-related scenes and names in the 2nd and 3rd rows, MACE, Receler, and AdvUnlearn generate corresponding explicit content. 
However, {\ours} effectively blocks the risk of generating unsafe content while maintaining the overall semantic consistency with nudity-related prompts.
The 4th row shows the preservation ability of different methods to common concept. It can be observed that SLD-Max, ESD, RACE, MACE, and AdvUnlearn all partially forget the concepts of kitchen or refrigerator, leading to the absence of the related objects in the generated images. In contrast, SAGE still maintains the semantic consistency on common concept prompt.
These visual differences demonstrate that {\ours} effectively erases nudity-related concepts while maintaining unrelated concepts, ensuring the safe generation.

\noindent \textbf{Artistic Style Erasure.}
The first two rows of Fig.~\ref{fig:vangogh} show the generated images for erasing the target Van Gogh style, while the last two rows show the images for other non-target artistic styles. It can be observed that SLD-Max and Receler, while erasing the Van Gogh style, also cause strong forgetting of other non-target styles. When confronted with the term \texttt{Starry Night}, which is strongly associated with Van Gogh's art concept, UCE, MACE, and RECE fail to erase the Van Gogh style. 
Meanwhile, ESD, RACE, and AdvUnlearn forget some patterns of other artistic styles, leading to significant perceptual shifts in the generated images compared to the original SD v1.4 on non-target artistic styles. \eg, Marilyn Monroe is transformed into a male in the 3rd row, and the Monet style is transformed from Impressionism to realistic photographic style in the 4th row. In comparison, our {\ours} not only effectively unlearns target Van Gogh style but also preserves the generation quality of other non-target styles as possible.

\subsection{Ablation Study}
\label{sec:exp_abl}
\noindent \textbf{Ablation about Components.}
To study the impact of each component, we conduct ablation studies on erasing nudity concept task and present results in Table~\ref{tab:ablation}.
The high RER and low ASR of Config. 1 and 2 indicate that only training with attack prompts allows the DM to effectively erase nudity concept. However, the extremely high FID and extremely low CLIP-S also mean the erased DM suffers a significant generation degradation for common concepts.
Comparing Config. 3 with 4, we observe that adding either $\mathcal{L}_{\mathrm{image}}$ or $\mathcal{L}_{\mathrm{graph}}$ improves the retain ability to common concepts. $\mathcal{L}_{\mathrm{image}}$ is more effective in maintaining local-level image generation quality (FID) but has a limited effect on maintaining global-level semantic alignment (CLIP-S). $\mathcal{L}_{\mathrm{graph}}$ can simultaneously maintain both generation quality and semantic alignment, but may slightly reduce the erasure effect.
Config. 5 shows that warm-up training slightly improves the erasure effect. It may be because the warm-up phase initially directs DM to focus on erasing concept words, which helps the subsequent exploration of the concept domain boundaries. 
Our {\ours} effectively unlearns unsafe concepts through semantic-augment erasure and mitigates the generation degradation caused by over-erasure through global-local collaborative retention, thereby achieving better comprehensive performance.

\begin{table}[tbp]
\caption{Ablation study on erasing nudity concept. $\tau_p$: attack prompts. \textit{wu}: warm-up training. $\mathcal{L}_{\mathrm{g}}$: semantic graph consistency loss. $\mathcal{L}_{\mathrm{i}}$: noise prediction consistency loss.}
	\begin{center}
 \setlength{\tabcolsep}{3pt}
		\resizebox{0.48\textwidth}{!}
  {
			\begin{tabular}{l@{\hskip -0.5pt}cccc|c|c@{\hskip -0.5pt}c|c|c}
				\Xhline{1pt}
 \multirow{2}{*}{\textbf{Config}} &\multicolumn{4}{c|}{\textbf{Components}} 
    &\multirow{2}{*}{\begin{tabular}[c]{@{}c@{}}\textbf{RER}\\($\uparrow$)\end{tabular}} 
    &\multirow{2}{*}{\begin{tabular}[c]{@{}c@{}}\textbf{FID}\\($\downarrow$)\end{tabular}} 
    &\multirow{2}{*}{\begin{tabular}[c]{@{}c@{}}\textbf{CLIP-S}\\($\uparrow$)\end{tabular}}
    &\multirow{2}{*}{\begin{tabular}[c]{@{}c@{}}\textbf{ASR}\\($\downarrow$)\end{tabular}}
    &\multirow{2}{*}{\begin{tabular}[c]{@{}c@{}}\cellcolor{mypink}{$~~\textbf{H}_\textbf{o}~~$}\\\cellcolor{mypink}(\text{$\uparrow$}) \end{tabular}}    
    \\
				 & $\tau_p$ & $\textit{wu}$ & $\mathcal{L}_{\mathrm{g}}$ & $\mathcal{L}_{\mathrm{i}}$ & & & &  &\\
				\Xhline{0.5pt}
                    SD v1.4 & - & - & - & - & 0.00 & 16.64 & 31.09 &- &\cellcolor{mypink}- \\
                    ESD & $\times$ & $\times$   & $\times$ & $\times$ &88.77 &18.18 &30.17 &40.84 &\cellcolor{mypink}84.21 \\
                    \Xhline{0.5pt}
                    1 & \checkmark & $\times$    & $\times$ & $\times$ &100.00 &73.38 &15.48 &0.00 &\cellcolor{mypink}68.11 \\
                    2 & \checkmark & \checkmark   & $\times$ & $\times$ &99.34 &66.39  &15.34 &0.00  &\cellcolor{mypink}68.43 \\
                    3 & \checkmark & \checkmark   & $\times$ & \checkmark  &97.02 &19.63  &25.99 &2.81  &\cellcolor{mypink}90.64 \\
                    4 & \checkmark & \checkmark   & \checkmark  & $\times$  &93.06  &18.24 &29.04 &5.63 &\cellcolor{mypink}93.01  \\
                    5 & \checkmark & $\times$  & \checkmark  & \checkmark  &96.70 &19.46  &29.31  &2.81 &\cellcolor{mypink}93.69 \\
                    \Xhline{0.5pt}
                    {\ours} & \checkmark & \checkmark   & \checkmark  & \checkmark  &98.01 &19.21  &29.53  &2.81  &\cellcolor{mypink}\textbf{94.28} \\
				\Xhline{1pt}
		\end{tabular}}
	\end{center}
	\label{tab:ablation}
\end{table}

\begin{table}[tbp]
\caption{Ablation study on the design of $\mathcal{L}_{\mathrm{attack}}$. }
	\begin{center}
 \setlength{\tabcolsep}{5pt}
		% \resizebox{0.48\textwidth}{!}
  {
			\begin{tabular}{l|c|cc|c|c}
				\Xhline{1pt}
 \multirow{2}{*}{\textbf{$\mathcal{L}_{\mathrm{attack}}$}}
    &\multirow{2}{*}{\begin{tabular}[c]{@{}c@{}}\textbf{RER}\\($\uparrow$)\end{tabular}} 
    &\multirow{2}{*}{\begin{tabular}[c]{@{}c@{}}\textbf{FID}\\($\downarrow$)\end{tabular}} 
    &\multirow{2}{*}{\begin{tabular}[c]{@{}c@{}}\textbf{CLIP-S}\\($\uparrow$)\end{tabular}}
    &\multirow{2}{*}{\begin{tabular}[c]{@{}c@{}}\textbf{ASR}\\($\downarrow$)\end{tabular}}
    &\multirow{2}{*}{\begin{tabular}[c]{@{}c@{}}\cellcolor{mypink}{$~~\textbf{H}_\textbf{o}~~$}\\\cellcolor{mypink}(\text{$\uparrow$}) \end{tabular}}    
    \\
    & & & & & \\
    \Xhline{0.5pt}
        SD v1.4  & 0.00 & 16.64 & 31.09 &- &\cellcolor{mypink}- \\
        ESD  &88.77 &18.18 &30.17 &40.84 &\cellcolor{mypink}84.21 \\
        \Xhline{0.5pt}  
        $\mathcal{H}_{2}$  &85.81 &18.91  &30.09 &30.99  &\cellcolor{mypink}84.90\\
        $-\mathcal{H}_{1}$  &93.73 &19.77 &28.57 &7.04 &\cellcolor{mypink}90.69\\
        $-\mathcal{H}_{1} + \mathcal{H}_{2}$   &96.03 &19.51 &29.29 &2.11  &\cellcolor{mypink}93.35\\
        \Xhline{0.5pt}
        \rule{0pt}{1.1em} 
        $-\mathcal{H}_{1} + \frac{\mathcal{H}_{2}}{\mathcal{H}_{1}}$  &98.01 &19.21  &29.53  &2.81  &\cellcolor{mypink}\textbf{94.28} \\
    \Xhline{1pt}
    \end{tabular}}
	\end{center}
	\label{tab:ablation}
\end{table}

\noindent \textbf{Ablations about $\mathcal{L}_{\mathrm{attack}}$.}
To explore valuable attack prompts which are within the target concept domain but distant from the target concept anchor, $\mathcal{L}_{\mathrm{attack}}$ (Eq.\eqref{eq:h_attack}) is defined for attack prompt generation. The ablation experiments are conducted on the attack loss design, with results shown in Table \ref{tab:ablation}.
Optimizing only $\mathcal{H}_{2}$ pushes attack prompts randomly away from the concept anchor. It underperforms compared to ESD in both erasure and retention. This decline may arise from semantic drift in randomly directed prompts, leading to imprecise concept erasure. While this method improves robustness by approximately 10\%, which is likely due to diverse attack prompts enhancing anti-attack capability.
In contrast, optimizing only $\mathcal{H}_{1}$ strictly aligns attack prompts with the concept anchor, improving both erasure efficacy and model robustness. The gains in erasure stem from the concept anchor providing a consistent optimization target during adversarial training. However, this approach degrades retention performance, as excessive unidirectional erasure weakens the ability to preserve unrelated concepts.
Linear combination of $\mathcal{H}_{1}$ and $\mathcal{H}_{2}$ achieves superior performance, balancing erasure efficacy, retention, and robustness.  This suggests that introducing controlled variance is useful during the unidirectional erasure process toward the concept anchor.
The attack loss used in this work increases $\mathcal{H}_{1}$ while further decreasing $\mathcal{H}_{2}$, generating attack prompts that remain within the concept domain yet explore its boundaries. These prompts refine the erasure process, ultimately yielding the best overall performance.

\begin{table}[tbp]
\centering
\caption{ASR evaluation of different methods.} 
\setlength{\tabcolsep}{4pt}
\begin{tabular}{lcccc}
\Xhline{1pt}
Method
&\textbf{RAB}
&\textbf{MMA}
&\textbf{P4D}
&\textbf{UnlearnDiff}
\\
\Xhline{0.5pt}
Receler \cite{receler} &9.86 &17.9 &42.96 &42.96  \\
RECE \cite{RECE} &11.97 &19.6 &31.69 &38.73   \\
RACE \cite{RACE} &21.83 &0.70 &26.06 &24.65  \\
AdvUnlearn \cite{Advunlearn}&6.33  &0.60 &\textbf{5.63} &\textbf{5.63} \\  
\Xhline{0.5pt}
{\ours}  &\textbf{2.81} &\textbf{0.50} &9.15 &9.86   \\
\Xhline{1pt}
\end{tabular}
\label{tab:attack_rebuttal}
\end{table}
\begin{table}[!tbp]
\caption{Comparison of nudity detection before and after transfering the text encoder of SAGE.}
\centering
    \setlength\tabcolsep{3pt} 
    % \resizebox{0.48\textwidth}{!}
    {
\begin{tabular}{lcc|c}
    \Xhline{1pt}
     Model  & Origin & Transfer & Relative Ratio \\
    \Xhline{0.5pt}
    LCM Dreamshaper v7 & 777 & 105 &$\downarrow$ \textbf{86.49\%}\\
    Dreamlike Photoreal v2.0 & 420 & 14  &$\downarrow$ \textbf{96.67\%}\\
    Openjourney v4 & 162 & 7  &$\downarrow$ \textbf{95.68\%}\\
    SDXL v1.0 (base) & 140 & 71  &$\downarrow$ \textbf{49.29\%}\\
    \Xhline{1pt}
\end{tabular}
    }
    \label{tab:more_baseline}
\end{table}

\subsection{Extended Experiments}
\noindent \textbf{More Red-teaming Methods.}
In the domain of multimodal generative safety, an ongoing arms race persists between jailbreak attacks and defense methods. 
To evaluate nudity-erasure robustness from more dimensions, we employ diverse red-teaming methods to assess robustness-specific approaches, \eg, Recelerc \cite{receler}, RECE \cite{RECE}, RACE \cite{RACE}, and AdvUnlearn \cite{Advunlearn}. Red-teaming methods include black-box attack methods such as Ring-A-Bell (RAB) \cite{RingABell} and MMA-Diffusion (MMA) \cite{mma}, which cannot access the parameters of victim models, and white-box attack methods such as Prompting4Debugging (P4D) \cite{prompting4debugging} and UnlearnDiff \cite{zhang2023generate}, which leverage gradient information and intermediate representations of victim models to optimize attacks precisely.
As shown in Table \ref{tab:attack_rebuttal}, our method {\ours} demonstrates strong robustness against both white-box and black-box attacks. Besides, it achieves the best anti-attack performance against black-box attacks, which have a wider application scope and lower deployment costs than white-box attacks.

\noindent \textbf{More Base Models.}
Unlike prior methods \cite{esd,RACE,UCE,MACE,receler,RECE} that modify the UNet, our approach updates the text encoder. This key difference allows our trained text encoder to be shared across multiple text-to-image models that use the same text encoder architecture, eliminating redeployment costs.
To quantitatively assess this zero-shot transfer capability, we evaluated our method on three Stable Diffusion v1.4 variants (LCM Dreamshaper v7 \cite{LCM}, Dreamlike Photoreal v2.0 \cite{dreamlike}, Openjourney v4 \cite{Openjourney}) and Stable Diffusion XL (SDXL) v1.0 \cite{sdxl}, which uses dual text encoders (OpenCLIP-ViT/G \cite{openclip} and CLIP-ViT/L \cite{clip}). 
We replace the CLIP-ViT/L text encoder in these models with our SAGE nudity-erasure model’s text encoder and measure the reduction in unsafe content generation on the I2P dataset.
As shown in Table \ref{tab:more_baseline}, our method reduces the probability of generating nudity by 86.49\%$\sim$96.67\% for single-text-encoder models. For SDXL, replacing only one text encoder still achieves a 49.29\% reduction. This training-free cross-model transfer capability indicates that our method has flexible adaptability in practical applications.

\section{Conclusion}
\label{sec:conclusion}
We propose a novel concept erasing method {\ours}. It breaks the convention of modeling concepts as fixed words and achieves the generalized concept domain erasing by the iterative self-check and self-erasure. 
Meanwhile, global-local collaborative retention provides dual protection mechanism for non-target concepts to ensure the model usability.
Extensive experiments demonstrate that {\ours} effectively and efficiently unlearns target concepts while maintaining high-quality image generation and semantic alignment.

\vspace{1mm}
\noindent\textbf{Data Availability Statements.} 
The authors declare that the data supporting the experiments in this study are available within the paper. The code will be available at \texttt{https://github.com/KevinLight831/SAGE}.


% \clearpage
\bibliographystyle{spmpsci}
\bibliography{sample}% common bib file

\begin{thebibliography}{10}
\providecommand{\url}[1]{{#1}}
\providecommand{\urlprefix}{URL }
\expandafter\ifx\csname urlstyle\endcsname\relax
  \providecommand{\doi}[1]{DOI~\discretionary{}{}{}#1}\else
  \providecommand{\doi}{DOI~\discretionary{}{}{}\begingroup \urlstyle{rm}\Url}\fi

\bibitem{nudenet}
Bedapudi, P.: Nudenet: Neural nets for nudity classification, detection and selective censoring (2019)

\bibitem{openclip}
Cherti, M., Beaumont, R., Wightman, R., Wortsman, M., Ilharco, G., Gordon, C., Schuhmann, C., Schmidt, L., Jitsev, J.: Reproducible scaling laws for contrastive language-image learning.
\newblock In: Proceedings of the IEEE/CVF Conference on Computer Vision and Pattern Recognition, pp. 2818--2829 (2023)

\bibitem{prompting4debugging}
Chin, Z.Y., Jiang, C.M., Huang, C.C., Chen, P.Y., Chiu, W.C.: Prompting4debugging: Red-teaming text-to-image diffusion models by finding problematic prompts.
\newblock In: International Conference on Machine Learning (ICML) (2024)

\bibitem{Hayden2024}
CNBC: Microsoft {AI} engineer says {Copilot} designer creates 'disturbing' images (2024).
\newblock Accessed: 2024-08-15

\bibitem{dhariwal2021diffusion}
Dhariwal, P., Nichol, A.: Diffusion models beat gans on image synthesis.
\newblock Advances in Neural Information Processing Systems \textbf{34}, 8780--8794 (2021)

\bibitem{ding2022cogview2}
Ding, M., Zheng, W., Hong, W., Tang, J.: Cogview2: Faster and better text-to-image generation via hierarchical transformers.
\newblock Advances in Neural Information Processing Systems \textbf{35}, 16890--16902 (2022)

\bibitem{dreamlike}
dreamlike.art: Dreamlike photoreal v2.0.
\newblock \url{https://huggingface.co/dreamlike-art/dreamlike-photoreal-2.0} (2023)

\bibitem{sd3}
Esser, P., Kulal, S., Blattmann, A., Entezari, R., M{\"u}ller, J., Saini, H., Levi, Y., Lorenz, D., Sauer, A., Boesel, F., et~al.: Scaling rectified flow transformers for high-resolution image synthesis.
\newblock In: Forty-first International Conference on Machine Learning (2024)

\bibitem{vqgan}
Esser, P., Rombach, R., Ommer, B.: Taming transformers for high-resolution image synthesis.
\newblock In: Proceedings of the IEEE/CVF conference on computer vision and pattern recognition, pp. 12873--12883 (2021)

\bibitem{gal2022image}
Gal, R., Alaluf, Y., Atzmon, Y., Patashnik, O., Bermano, A.H., Chechik, G., Cohen-Or, D.: An image is worth one word: Personalizing text-to-image generation using textual inversion.
\newblock arXiv preprint arXiv:2208.01618  (2022)

\bibitem{esd}
Gandikota, R., Materzynska, J., Fiotto-Kaufman, J., Bau, D.: Erasing concepts from diffusion models.
\newblock In: Proceedings of the IEEE/CVF International Conference on Computer Vision, pp. 2426--2436 (2023)

\bibitem{UCE}
Gandikota, R., Orgad, H., Belinkov, Y., Materzy{\'n}ska, J., Bau, D.: Unified concept editing in diffusion models.
\newblock In: Proceedings of the IEEE/CVF Winter Conference on Applications of Computer Vision, pp. 5111--5120 (2024)

\bibitem{RECE}
Gong, C., Chen, K., Wei, Z., Chen, J., Jiang, Y.G.: Reliable and efficient concept erasure of text-to-image diffusion models.
\newblock In: European Conference on Computer Vision, pp. 73--88. Springer (2024)

\bibitem{animatediff}
Guo, Y., Yang, C., Rao, A., Liang, Z., Wang, Y., Qiao, Y., Agrawala, M., Lin, D., Dai, B.: Animatediff: Animate your personalized text-to-image diffusion models without specific tuning.
\newblock arXiv preprint arXiv:2307.04725  (2023)

\bibitem{Openjourney}
Hero, P.: Openjourney v4.
\newblock \url{https://huggingface.co/prompthero/openjourney-v4} (2023)

\bibitem{clipscore}
Hessel, J., Holtzman, A., Forbes, M., Bras, R.L., Choi, Y.: Clipscore: A reference-free evaluation metric for image captioning.
\newblock arXiv preprint arXiv:2104.08718  (2021)

\bibitem{FID}
Heusel, M., Ramsauer, H., Unterthiner, T., Nessler, B., Hochreiter, S.: Gans trained by a two time-scale update rule converge to a local nash equilibrium.
\newblock Advances in neural information processing systems \textbf{30} (2017)

\bibitem{ddpm}
Ho, J., Jain, A., Abbeel, P.: Denoising diffusion probabilistic models.
\newblock Advances in neural information processing systems \textbf{33}, 6840--6851 (2020)

\bibitem{ho2021classifier}
Ho, J., Salimans, T.: Classifier-free diffusion guidance.
\newblock In: NeurIPS 2021 Workshop on Deep Generative Models and Downstream Applications (2021)

\bibitem{receler}
Huang, C.P., Chang, K.P., Tsai, C.T., Lai, Y.H., Yang, F.E., Wang, Y.C.F.: Receler: Reliable concept erasing of text-to-image diffusion models via lightweight erasers.
\newblock In: European Conference on Computer Vision, pp. 360--376. Springer (2024)

\bibitem{jiang2023ai}
Jiang, H.H., Brown, L., Cheng, J., Khan, M., Gupta, A., Workman, D., Hanna, A., Flowers, J., Gebru, T.: Ai art and its impact on artists.
\newblock In: Proceedings of the 2023 AAAI/ACM Conference on AI, Ethics, and Society, pp. 363--374 (2023)

\bibitem{RACE}
Kim, C., Min, K., Yang, Y.: Race: Robust adversarial concept erasure for secure text-to-image diffusion model.
\newblock In: European Conference on Computer Vision, pp. 461--478. Springer (2024)

\bibitem{sdd}
Kim, S., Jung, S., Kim, B., Choi, M., Shin, J., Lee, J.: Towards safe self-distillation of internet-scale text-to-image diffusion models.
\newblock arXiv preprint arXiv:2307.05977  (2023)

\bibitem{kingma2013auto}
Kingma, D.P.: Auto-encoding variational bayes.
\newblock arXiv preprint arXiv:1312.6114  (2013)

\bibitem{AC}
Kumari, N., Zhang, B., Wang, S.Y., Shechtman, E., Zhang, R., Zhu, J.Y.: Ablating concepts in text-to-image diffusion models.
\newblock In: Proceedings of the IEEE/CVF International Conference on Computer Vision, pp. 22691--22702 (2023)

\bibitem{coco}
Lin, T.Y., Maire, M., Belongie, S., Hays, J., Perona, P., Ramanan, D., Doll{\'a}r, P., Zitnick, C.L.: Microsoft coco: Common objects in context.
\newblock In: Computer Vision--ECCV 2014: 13th European Conference, Zurich, Switzerland, September 6-12, 2014, Proceedings, Part V 13, pp. 740--755. Springer (2014)

\bibitem{liu2023grounding}
Liu, S., Zeng, Z., Ren, T., Li, F., Zhang, H., Yang, J., Li, C., Yang, J., Su, H., Zhu, J., et~al.: Grounding dino: Marrying dino with grounded pre-training for open-set object detection.
\newblock arXiv preprint arXiv:2303.05499  (2023)

\bibitem{MACE}
Lu, S., Wang, Z., Li, L., Liu, Y., Kong, A.W.K.: Mace: Mass concept erasure in diffusion models.
\newblock In: Proceedings of the IEEE/CVF Conference on Computer Vision and Pattern Recognition, pp. 6430--6440 (2024)

\bibitem{LCM}
Luo, S., Tan, Y., Huang, L., Li, J., Zhao, H.: Latent consistency models: Synthesizing high-resolution images with few-step inference.
\newblock arXiv preprint arXiv:2310.04378  (2023)

\bibitem{SPM}
Lyu, M., Yang, Y., Hong, H., Chen, H., Jin, X., He, Y., Xue, H., Han, J., Ding, G.: One-dimensional adapter to rule them all: Concepts diffusion models and erasing applications.
\newblock In: Proceedings of the IEEE/CVF Conference on Computer Vision and Pattern Recognition, pp. 7559--7568 (2024)

\bibitem{PGD}
Madry, A., Makelov, A., Schmidt, L., Tsipras, D., Vladu, A.: Towards deep learning models resistant to adversarial attacks.
\newblock In: International Conference on Learning Representations (2018)

\bibitem{midjourney}
MidJourney, I.: Midjourney.
\newblock \url{https://www.midjourney.com} (2023).
\newblock V5

\bibitem{dalle2}
OpenAI: Dall·e 2.
\newblock \url{https://openai.com/dall-e-2} (2022).
\newblock V1

\bibitem{gpt2023}
OpenAI: Gpt-4 technical report  (2023)

\bibitem{connor2022sd}
O’Connor, R.: Stable diffusion 1 vs 2 - what you need to know (2022)

\bibitem{sdxl}
Podell, D., English, Z., Lacey, K., Blattmann, A., Dockhorn, T., M{\"u}ller, J., Penna, J., Rombach, R.: Sdxl: Improving latent diffusion models for high-resolution image synthesis.
\newblock arXiv preprint arXiv:2307.01952  (2023)

\bibitem{tatum2023porn}
Post, T.W.: Ai porn and deepfakes are a growing threat to women’s consent (2023).
\newblock Accessed: 2024-08-15

\bibitem{qu2023unsafe}
Qu, Y., Shen, X., He, X., Backes, M., Zannettou, S., Zhang, Y.: Unsafe diffusion: On the generation of unsafe images and hateful memes from text-to-image models.
\newblock In: Proceedings of the 2023 ACM SIGSAC Conference on Computer and Communications Security, pp. 3403--3417 (2023)

\bibitem{clip}
Radford, A., Kim, J.W., Hallacy, C., Ramesh, A., Goh, G., Agarwal, S., Sastry, G., Askell, A., Mishkin, P., Clark, J., et~al.: Learning transferable visual models from natural language supervision.
\newblock In: International conference on machine learning, pp. 8748--8763. PMLR (2021)

\bibitem{ramesh2022hierarchical}
Ramesh, A., Dhariwal, P., Nichol, A., Chu, C., Chen, M.: Hierarchical text-conditional image generation with clip latents.
\newblock arXiv preprint arXiv:2204.06125 \textbf{1}(2), 3 (2022)

\bibitem{rando2022red}
Rando, J., Paleka, D., Lindner, D., Heim, L., Tram{\`e}r, F.: Red-teaming the stable diffusion safety filter.
\newblock arXiv preprint arXiv:2210.04610  (2022)

\bibitem{rombach2022sd2}
Rombach, R.: Stable diffusion 2.0 release  (2022)

\bibitem{SD1_4}
Rombach, R., Blattmann, A., Lorenz, D., Esser, P., Ommer, B.: High-resolution image synthesis with latent diffusion models.
\newblock In: Proceedings of the IEEE/CVF conference on computer vision and pattern recognition, pp. 10684--10695 (2022)

\bibitem{ronneberger2015u}
Ronneberger, O., Fischer, P., Brox, T.: U-net: Convolutional networks for biomedical image segmentation.
\newblock In: Medical image computing and computer-assisted intervention--MICCAI 2015: 18th international conference, Munich, Germany, October 5-9, 2015, proceedings, part III 18, pp. 234--241. Springer (2015)

\bibitem{roose2022art}
Roose, K.: An ai-generated picture won an art prize. artists aren’t happy  (2022)

\bibitem{SLD}
Schramowski, P., Brack, M., Deiseroth, B., Kersting, K.: Safe latent diffusion: Mitigating inappropriate degeneration in diffusion models.
\newblock In: Proceedings of the IEEE/CVF Conference on Computer Vision and Pattern Recognition, pp. 22522--22531 (2023)

\bibitem{q16}
Schramowski, P., Tauchmann, C., Kersting, K.: Can machines help us answering question 16 in datasheets, and in turn reflecting on inappropriate content?
\newblock In: Proceedings of the 2022 ACM Conference on Fairness, Accountability, and Transparency, pp. 1350--1361 (2022)

\bibitem{schuhmann2022laion}
Schuhmann, C., Beaumont, R., Vencu, R., Gordon, C., Wightman, R., Cherti, M., Coombes, T., Katta, A., Mullis, C., Wortsman, M., et~al.: Laion-5b: An open large-scale dataset for training next generation image-text models.
\newblock Advances in Neural Information Processing Systems \textbf{35}, 25278--25294 (2022)

\bibitem{setty2023suit}
Setty, R.: Ai art generators hit with copyright suit over artists’ images (2023)

\bibitem{SmithMano2022}
SmithMano: Tutorial: How to remove the safety filter in 5 seconds (2022)

\bibitem{song2020denoising}
Song, J., Meng, C., Ermon, S.: Denoising diffusion implicit models.
\newblock arXiv preprint arXiv:2010.02502  (2020)

\bibitem{RingABell}
Tsai, Y.L., Hsu, C.Y., Xie, C., Lin, C.H., Chen, J.Y., Li, B., Chen, P.Y., Yu, C.M., Huang, C.Y.: Ring-a-bell! how reliable are concept removal methods for diffusion models?
\newblock In: The Twelfth International Conference on Learning Representations (2024)

\bibitem{sv3d}
Voleti, V., Yao, C.H., Boss, M., Letts, A., Pankratz, D., Tochilkin, D., Laforte, C., Rombach, R., Jampani, V.: Sv3d: Novel multi-view synthesis and 3d generation from a single image using latent video diffusion.
\newblock In: European Conference on Computer Vision, pp. 439--457. Springer (2025)

\bibitem{mma}
Yang, Y., Gao, R., Wang, X., Ho, T.Y., Xu, N., Xu, Q.: Mma-diffusion: Multimodal attack on diffusion models.
\newblock In: Proceedings of the IEEE/CVF Conference on Computer Vision and Pattern Recognition, pp. 7737--7746 (2024)

\bibitem{sneakyprompt}
Yang, Y., Hui, B., Yuan, H., Gong, N., Cao, Y.: Sneakyprompt: Jailbreaking text-to-image generative models.
\newblock arXiv preprint arXiv:2305.12082  (2023)

\bibitem{zhang2024adversarial}
Zhang, C., Hu, M., Li, W., Wang, L.: Adversarial attacks and defenses on text-to-image diffusion models: A survey.
\newblock Information Fusion p. 102701 (2024)

\bibitem{FMN}
Zhang, G., Wang, K., Xu, X., Wang, Z., Shi, H.: Forget-me-not: Learning to forget in text-to-image diffusion models.
\newblock In: Proceedings of the IEEE/CVF Conference on Computer Vision and Pattern Recognition, pp. 1755--1764 (2024)

\bibitem{Advunlearn}
Zhang, Y., Chen, X., Jia, J., Zhang, Y., Fan, C., Liu, J., Hong, M., Ding, K., Liu, S.: Defensive unlearning with adversarial training for robust concept erasure in diffusion models.
\newblock arXiv preprint arXiv:2405.15234  (2024)

\bibitem{zhang2023generate}
Zhang, Y., Jia, J., Chen, X., Chen, A., Zhang, Y., Liu, J., Ding, K., Liu, S.: To generate or not? safety-driven unlearned diffusion models are still easy to generate unsafe images... for now.
\newblock ECCV  (2024)

\bibitem{zhang2023large}
Zhang, Z., Fang, M., Chen, L., Namazi-Rad, M.R., Wang, J.: How do large language models capture the ever-changing world knowledge? a review of recent advances.
\newblock In: Proceedings of the 2023 Conference on Empirical Methods in Natural Language Processing, pp. 8289--8311 (2023)

\end{thebibliography}

\end{document}